\documentclass[conference]{IEEEtran}
\usepackage{amsmath}
\usepackage{amssymb}
\usepackage{hyperref}
\usepackage{graphicx}

\usepackage{listings}

\usepackage{url} 

\usepackage{natbib}
\setcitestyle{square,numbers}

\usepackage{float}
\usepackage{tabularx}
\usepackage{makecell}
\usepackage{array}

\usepackage{fancyhdr}  
\title{KERAIA: An Adaptive and Explainable Framework for Dynamic Knowledge Representation and Reasoning}

\author{
    Stephen Richard Varey, Alessandro Di Stefano, The Anh Han \\
    Teesside University, Middlesbrough, UK \\
    Email: \{S.Varey, A.DiStefano, T.Han\}@tees.ac.uk
}

\begin{document}

\maketitle

\begin{abstract}
In this paper, we introduce KERAIA, a novel framework and software platform for symbolic knowledge engineering designed to address the persistent challenges of representing, reasoning with, and executing knowledge in dynamic, complex, and context-sensitive environments. The central research question that motivates this work is: How can unstructured, often tacit, human expertise be effectively transformed into computationally tractable algorithms that AI systems can efficiently utilise? KERAIA seeks to bridge this gap by building on foundational concepts such as Minsky's frame-based reasoning and K-lines, while introducing significant innovations. These include Clouds of Knowledge for dynamic aggregation, Dynamic Relations (DRels) for context-sensitive inheritance, explicit Lines of Thought (LoTs) for traceable reasoning, and Cloud Elaboration for adaptive knowledge transformation. This approach moves beyond the limitations of traditional, often static, knowledge representation paradigms. KERAIA is designed with Explainable AI (XAI) as a core principle, ensuring transparency and interpretability, particularly through the use of LoTs. The paper details the framework's architecture, the KSYNTH representation language, and the General Purpose Paradigm Builder (GPPB) to integrate diverse inference methods within a unified structure. We validate KERAIA's versatility, expressiveness, and practical applicability through detailed analysis of multiple case studies spanning naval warfare simulation, industrial diagnostics in water treatment plants, and strategic decision-making in the game of RISK. Furthermore, we provide a comparative analysis against established knowledge representation paradigms (including ontologies, rule-based systems, and knowledge graphs) and discuss the implementation aspects and computational considerations of the KERAIA platform.
\end{abstract}


\section{Introduction}
\label{introduction}

Artificial intelligence (AI) research has long pursued the goal of emulating human cognitive capabilities, particularly the remarkable capacity for adaptive reasoning and decision making in complex, uncertain, and dynamically changing environments \citep{marcus2020, Wooldridge2024}. A fundamental prerequisite for achieving this intelligence lies in effective knowledge representation (KR), the study of how information about the world can be structured and encoded such that machines can interpret, manipulate, and reason with it effectively \citep{brachman1985, sowa1999}. Despite decades of progress, a significant chasm persists between the richness, fluidity, and often tacit nature of human expertise and the structured, computationally tractable formats required by AI algorithms. This challenge is particularly acute in domains characterized by high complexity, dynamism, and the need for explainable reasoning, such as military command and control, industrial process monitoring, medical diagnostics, financial risk assessment, and strategic planning.

Traditional KR paradigms, while foundational, often exhibit significant limitations when faced with the demands of these real-world scenarios. Semantic networks \citep{sowa1992semantic}, frame systems \citep{minsky1980}, rule-based expert systems \citep{buchanan2005}, logic-based approaches that include description logics (DL) that underpin ontologies such as RDF/OWL \citep{baader2017, hitzler2009owl3}, and case-based reasoning (CBR) \citep{aamodt1994cbr} have all contributed valuable concepts. However, they often struggle with the following.
\begin{itemize}
    \item \textbf{Brittleness and Rigidity:} Static hierarchies and predefined rules often fail to adapt gracefully to unforeseen situations or exceptions, leading to system failures or incorrect conclusions \citep{levesque1986}.
    \item \textbf{Lack of Context Sensitivity:} Many traditional systems lack robust mechanisms to modulate knowledge representation or inference based on the current context, limiting their ability to apply knowledge appropriately in varying situations.
    \item \textbf{Difficulty Handling Real-Time Updates:} Incorporating new information or adapting knowledge structures dynamically in response to rapidly changing environments can be computationally expensive or architecturally challenging for many existing frameworks.
    \item \textbf{The Explainability Gap:} Reasoning processes within complex logical systems, large rule bases, or opaque machine learning models can be difficult or impossible for human users to understand, hindering trust, verification, and adoption in critical applications where transparency is paramount \citep{miller2019, gunning2017explainable}.
\end{itemize}
These limitations echo the challenges observed in ambitious historical projects like Japan’s Fifth Generation Computer Systems, which struggled partly due to an overemphasis on formal logic models that did not align well with the fluid demands of practical applications and market needs.

To address these persistent challenges, this paper introduces KERAIA (Knowledge Engineering and Reference AI Architecture), a novel framework and accompanying software platform for symbolic knowledge engineering \citep{varey2024keraia,vareySKIMA2025}. KERAIA is designed specifically to bridge the gap between unstructured human expertise and executable AI algorithms, focusing on adaptability, context-sensitivity, and explainability. It revisits and extends foundational ideas, notably Minsky's concepts of frames and K lines (representing memory activation patterns) \citep{minsky1988}, integrating them with several conceptual and architectural innovations derived from extensive research detailed in \citep{varey2024keraia,vareySKIMA2025}. The core research question driving KERAIA's development is: \textit{How can unstructured human expertise be transformed into algorithms that AI can efficiently utilize in dynamic and complex environments while ensuring the reasoning process is transparent and traceable?}

Central to the KERAIA framework are the following key innovations.
\begin{itemize}
    \item \textbf{Clouds of Knowledge:} Moving beyond rigid hierarchies, clouds represent dynamic, context-driven aggregations of knowledge sources (KSs). They can encapsulate specific domains, viewpoints, scenarios, or even hypothetical states, and can be nested recursively, reflecting the multifaceted and layered nature of human mental models and complex systems.
    \item \textbf{Dynamic Relations (DRels):} Offering a significant departure from static inheritance (e.g., `is-a` links in traditional ontologies), DRels provide a flexible, context-sensitive mechanism for property and method sharing between KSs. Inheritance becomes conditional, dynamically evaluated based on factors such as spatial proximity, temporal relevance, or specific states, allowing knowledge to adapt situationally rather than being restricted by fixed classifications.
    \item \textbf{Knowledge Lines (KLines) and Lines of Thought (LoT):} Inspired by Minsky's K lines, these are first-class explicit constructs that represent traceable pathways for reasoning and activation of knowledge. LoTs connect sequences of KSs, potentially spanning multiple Clouds or Dimensions, to guide inference, structure complex processes, and provide a transparent audit trail for explainability (XAI).
    \item \textbf{Cloud Elaboration:} This mechanism facilitates the dynamic linking, transformation, and adaptation of knowledge within and between clouds. It allows the system to restructure its understanding, resolve conflicts, or generate new insights by applying transformation functions based on the current context or inference goals.
    \item \textbf{Integrated Software Platform:} KERAIA is not merely a conceptual model but is embodied in a robust software platform designed to support the entire lifecycle of knowledge engineering. This includes tools for collaborative knowledge acquisition (e.g., web-based interviews, Ultragraphs for visual modeling), the KSYNTH language to represent knowledge structures, a multiparadigm inference engine facilitated by the General Purpose Paradigm Builder (GPPB) and features for version control and collaborative refinement.
\end{itemize}

KERAIA aims to advance the state-of-the-art by offering a more flexible, dynamic, explainable, and integrated approach compared to many traditional KR systems. By providing a platform that supports collaborative knowledge acquisition, multi-paradigm reasoning, and transparent execution, the aim is to make sophisticated symbolic AI more practical and accessible for complex real-world applications.

The primary contributions of this work, significantly expanding upon the initial conference presentation \citep{varey2024keraia}, are:

\begin{enumerate}
    \item \textbf{Detailed Elaboration of the KERAIA framework:} A comprehensive description of the conceptual framework, its novel KR constructs (Clouds, DRels, LoTs, Cloud Elaboration, Forks, Junctures, Dimensions, etc.), the KSYNTH representation language, and the underlying software platform architecture, including the GPPB for integrating diverse inference methods.
    \item \textbf{Demonstration of Generalizability and Applicability:} Validation of KERAIA's versatility through in-depth analysis of multiple and diverse case studies: the original naval warfare scenario, diagnostics in water treatment plants, and strategic decision-making in the game of RISK.
    \item \textbf{Emphasis on Explainable AI (XAI):} A thorough discussion of how KERAIA achieves transparency and traceability, particularly through explicit modeling and use of Lines of Thought (LoTs), providing concrete examples.
    \item \textbf{Implementation Insights and Comparative Analysis:} Discussion of the implementation aspects of the KERAIA software platform, computational considerations, and a comparative analysis positioning KERAIA against other established KR paradigms (e.g., Ontologies + SWRL, CBR, KG, Rule-Based Systems).
\end{enumerate}

This paper is structured to provide a comprehensive understanding of KERAIA and its contributions. Section \ref{relatedwork} provides a review of related work in knowledge representation, critically examining the limitations of existing approaches, and positioning KERAIA within the broader AI landscape, explicitly comparing it with paradigms such as RDF/OWL. Section \ref{framework} delves  into the KERAIA framework, detailing its architecture, core concepts (including a dedicated subsection that elaborates on LoTs with examples), the KSYNTH language, and crucial implementation aspects such as the GPPB and the software platform. Section \ref{evaluation} presents an  evaluation of the framework, revisiting the naval scenario and detailing the water treatment and RISK case studies to demonstrate generalizability, alongside a discussion of benchmarking considerations and computational feasibility. Section \ref{discussion} offers a critical discussion of KERAIA's advantages, limitations, and implications, considering its potential impact and challenges. Finally, Section \ref{conclusion} summarizes the key contributions and outlines promising directions for future research.

\section{Background and Related Work}
\label{relatedwork}

Knowledge representation (KR) is a cornerstone of artificial intelligence, concerned with how knowledge can be symbolized and manipulated computationally to enable intelligent behavior \citep{brachman1985, sowa1999}. The field has produced a diverse array of paradigms over several decades, each offering different trade-offs in terms of expressiveness, computational tractability, ease of knowledge acquisition, and inherent reasoning capabilities. Understanding this landscape is crucial for positioning KERAIA and recognizing its specific contributions.

\subsection{Foundational Knowledge Representation Paradigms}

\begin{itemize}
    \item \textbf{Semantic Networks:} Among the earliest KR formalisms, semantic networks represent knowledge as graphs in which nodes correspond to concepts and edges represent semantic relationships between them \citep{Quillian1968}. Although intuitive, early versions often lacked formal semantics, leading to ambiguity in interpretation.
    \item \textbf{Frame Systems:} Introduced by Minsky \citep{minsky1974framework, minsky1980}, frame systems organize knowledge into data structures (frames) representing stereotypical situations or concepts. Frames contain slots for attributes, which can hold values, default values, constraints, or procedures (demons) triggered by access or updates. Frames support inheritance through hierarchical links (e.g., `is-a`, `part-of`), allowing properties to be shared. KERAIA draws inspiration from frames, but significantly extends the concepts of inheritance and dynamic organization.
    \item \textbf{Rule-Based Systems:} These systems encode knowledge primarily as conditional \textit{IF-THEN} rules (productions). An inference engine matches rules against facts in a working memory and executes the actions of triggered rules, enabling deductive reasoning. Expert systems of the 1980s relied heavily on this paradigm \citep{buchanan2005}. Although powerful in encoding heuristic knowledge, large rule bases can become difficult to manage, maintain, and verify for consistency, and they often struggle with representing complex structural relationships.
    \item \textbf{Logic-Based Approaches:} Formal logic, including propositional and first-order predicate calculus, provides a rigorous foundation for KR with well-defined syntax, semantics, and sound inference procedures. Description logics (DL) \citep{baader2017} are a family of logic-based formalisms designed to represent terminological knowledge. DLs form the theoretical basis for the Web Ontology Language (OWL) \citep{hitzler2009owl3}, a W3C standard to define ontologies on the Semantic Web. OWL, built upon the Resource Description Framework (RDF), allows defining classes (concepts), properties (roles), and individuals, along with axioms specifying relationships between them (e.g., subclass relationships, property restrictions). Automated reasoners can then infer implicit knowledge based on explicit ontology definitions.
    \item \textbf{Case-Based Reasoning (CBR):} CBR systems solve new problems by retrieving and adapting solutions from similar past problems (cases) stored in a case library \citep{aamodt1994cbr}. CBR emphasizes learning from experience and is effective in domains where acquiring explicit models is difficult but past examples are plentiful.
    \item \textbf{Knowledge Graphs (KGs):} Emerging from Semantic Web technologies and large-scale information extraction, KGs represent knowledge as graph structures, typically consisting of entities (nodes) and relationships (edges) between them, often stored as RDF triples (subject-predicate-object) \citep{hogan2021knowledge}. KGs excel at integrating heterogeneous data and supporting complex queries, but standard models like RDF/OWL can still face challenges in representing highly dynamic or context-dependent information.
\end{itemize}

\subsection{Cognitive Models and Their Influence on AI Architectures}

The journey to develop AI systems that can understand, learn, and reason like humans has been marked by an exploration of various cognitive principles. Marvin Minsky's pioneering work \citep{minsky1980} on "K-Lines" laid the groundwork for understanding the dynamic flow of reasoning and the interconnections between different knowledge units. Minsky proposed that human reasoning is not purely linear, but involves complex pathways that weave together disparate pieces of knowledge and memories. Subsequent research has delved into the roles of metaphors \citep{lakoff2008}, conceptual blending \citep{hiraga1999}, analogical reasoning \citep{hall1989,minervino2023,forbus2017extending}, and human cognition \citep{varela2017,schaffernicht2024}, which have all been instrumental in moving beyond purely symbolic representations toward more sophisticated models of human cognition.

In the realm of AI architectures, the work of \cite{bonasso1988} underscores the critical importance of "areas of interest" in achieving effective situational awareness. He argues that AI systems should prioritize relevant information rather than attempting to process all available data, especially in domains where real-time decision making is crucial. Early naval warfare models often used a layered architecture that included data fusion, situation assessment, and resource allocation \citep{young2015}. However, the increasing complexity of modern warfare scenarios, which now includes the domains of space and cyberspace, requires more sophisticated approaches \citep{Seng2018}.

\subsection{Challenges and Advances in Knowledge Engineering}

The rise of "big data" has further complicated the landscape of knowledge engineering.  \cite{Harrison2014} discusses the challenges of handling large data sets in real time, emphasizing the need for robust and scalable architectures. He argues that traditional approaches to data processing are often insufficient to manage the massive volume and velocity of data in modern environments. The development of 'digital twins' \citep{Zhou2020}, which create virtual representations of real-world systems, presents opportunities for AI solutions driven by knowledge \citep{Amenyo2018}. These advancements highlight the need for AI systems that can adapt to dynamic data flows, learn from massive datasets, and operate effectively in complex, interconnected environments.

Knowledge representation remains central to AI, serving as the bridge between raw data and meaningful insights.  \cite{brachman1985} provide a comprehensive overview of this field, emphasizing its fundamental role in enabling AI systems to process and reason about information. They argue that knowledge representation goes beyond just storing data; it involves capturing the structure and meaning of data, thereby allowing AI systems to perform complex reasoning tasks.  \cite{levesque1986} further emphasizes the importance of accurate and effective knowledge representation to enable AI systems to perform intelligent tasks, from natural language processing to decision making.

Knowledge graphs, which represent information as a network of nodes and edges, have emerged as powerful tools for knowledge representation \citep{neo2023,apache2023}. These graphs facilitate the representation of complex relationships between entities, enabling AI systems to perform tasks such as entity linking, question answering, and knowledge inference. Ji et al. \citep{ji2021} offer a comprehensive survey of knowledge graphs, discussing their advantages, such as handling uncertainty and providing interpretable results, and the challenges of construction and maintenance, including data quality issues and the need for scalable algorithms.

Traditional ontologies, with their structured hierarchies of classes and instances, have proven effective in organizing knowledge in specific domains \citep{sowa1999}. However, they often fall short in capturing the dynamic and multi-dimensional nature of real-world scenarios. Wildman \citep{wildman2010} critiques the limitations of traditional ontologies, arguing for more flexible and context-sensitive approaches to knowledge representation. He highlights the rigidity of traditional ontologies, which often fail to represent the complex and evolving relationships inherent in real-world contexts.

\subsection{The Ethics and Explainability in AI Systems}

The increasing reliance on AI systems in various domains raises critical questions about transparency, explainability, and ethical implications. Versailles et al. \cite{Versailles2021} explore the intersection of argumentation and explainable AI, stressing the need for AI systems to provide clear and justifiable explanations for their decisions. They argue that explainability is crucial for building trust and understanding in AI systems, allowing users to make informed decisions and to hold AI responsible for its actions.  \cite{sison2023} discuss the ethical challenges associated with emerging AI technologies, particularly those involving large language models like ChatGPT \citep{haleem2022}. They emphasize the importance of responsible AI development and deployment, ensuring that AI systems are used ethically and do not perpetuate biases or cause harm.

The growing focus on explainable AI is driven by the need for trust and accountability in AI systems, particularly in critical domains such as healthcare, finance, and law enforcement, where decisions can have significant consequences. \cite{rafferty2017} highlight the importance of intention recognition in assisted living within smart homes and the use of intention recognition for knowledge representation and reasoning in assisted living \citep{TheAnh20011,HanBook2013}. They argue that AI systems in these settings should be capable of understanding and responding to human intentions, leading to more personalized and effective assistive care.

\subsection{Benchmarking in AI}

As AI systems become increasingly sophisticated, the need for robust benchmarking and evaluation methodologies becomes more critical. Traditional metrics often assess performance in specific tasks, but may fail to capture the broader impact of AI systems on aspects such as explainability, fairness, and ethical considerations. The question of whether users achieve a pragmatic understanding of AI systems through explanations is central to evaluating explainable AI (XAI). Studies have focused on assessing the quality of the explanations provided by AI systems \citep{doshivelez2017}, the satisfaction of users with these explanations \citep{hoffman2018}, and the extent to which users understand AI systems through various comprehension tests \citep{miller2019}. Furthermore, research has explored the role of curiosity in motivating users to seek explanations \citep{berlyne1960}, the appropriateness of trust and reliance on AI systems \citep{langer2020}, and the performance of human-XAI work systems in supporting collaborative work \citep{kaur2020}.

\subsection{Limitations of Traditional Paradigms in Dynamic Contexts}

While these paradigms have proven successful in various applications, they often encounter difficulties when applied to domains characterized by dynamism, complexity, and context-dependency. Key limitations include:

\begin{itemize}
    \item \textbf{Static Structures:} Ontologies, frame hierarchies, and rule sets are often designed based on a static view of the domain. Adapting these structures in real-time to reflect changing situations or new information can be cumbersome or computationally prohibitive.
    \item \textbf{Limited Contextual Reasoning:} Many traditional KR formalisms lack explicit mechanisms to represent and reason about context. This makes it difficult to ensure that knowledge is applied appropriately based on the current situation, leading to potentially irrelevant or incorrect inferences \citep{brezillon2008context, acar2020reasoning, benerecetti2000contextual, schmidtke2012contextual, bouquet2003theories}.
    \item \textbf{Rigid Inheritance:} Standard inheritance mechanisms (e.g., `is-a` links) are typically static and monotonic. They cannot easily capture situations where relationships or properties change based on context, such as an object temporarily inheriting properties from its container or location.
    \item \textbf{Integration Challenges:} Integrating knowledge represented in different formalisms (e.g. rules, ontologies, procedural code) within a single system can be complex, often requiring ad hoc bridges or middleware.
    \item \textbf{Knowledge Acquisition Bottleneck:} Eliciting, formalizing and maintaining knowledge remains a significant challenge (the 
\end{itemize}

\begin{itemize}
    \item \textbf{Clouds of Knowledge:} Representing a fundamental shift from static hierarchies, Clouds are dynamic, potentially nested, self-contained knowledge spaces. A Cloud aggregates Knowledge Sources (KSs) contextually, representing a specific domain (e.g., `SensorFusionCloud`), scenario (`HighThreatScenarioCloud`), viewpoint (`FriendlyForcesViewCloud`), or even a hypothetical state (`WhatIfEngineFailureCloud`). Their ability to evolve and contain other Clouds allows for a recursive, multi-faceted representation mirroring the complexity of real-world knowledge and human thought. Clouds provide scope and context for the KSs they contain.
    \item \textbf{Knowledge Sources (KS):} These are the atomic building blocks within Clouds. A KS is analogous to an enhanced frame or object, encapsulating not only data (slots/fillers, potentially nested to arbitrary depths using KSYNTH) but also associated inference methods (`responders`), event triggers (`attractors`), activation signals (`impulses`), and state change notifications (`pulses`). Each KS acts as a unit of information and localized reasoning, adaptable in real-time based on context. KSs can represent entities (e.g., `Ship`, `Sensor`), concepts (`ThreatLevel`), processes (`TargetTracking`), or abstract relationships.
    \item \textbf{Dynamic Relations (DRels):} DRels are a cornerstone of KERAIA’s flexibility, providing a context-sensitive inheritance and relationship mechanism. Unlike the rigid ‘is-a’ hierarchies of traditional ontologies (e.g., `rdfs:subClassOf`), DRels allow properties, methods (`responders`), and even structural components to be shared or linked between KSs based on dynamically evaluated conditions. These conditions can involve spatial proximity (`distance < 10km`), temporal relevance (`timeSinceLastUpdate < 5min`), specific states (`status == active`), or complex logical expressions evaluated at runtime. This enables adaptive knowledge sharing that reflects situational nuances rather than fixed classifications. For example, a `FighterJet` KS might dynamically inherit targeting parameters from a `CommandCenter` KS only when within a specific operational zone and authorized, conditions checked via DRels.
    \item \textbf{Cloud Elaboration:} This is the process by which knowledge within Clouds is dynamically linked, transformed, and adapted based on situational needs or inference goals. It involves applying transformation functions (often associated with DRels or specific KS responders) to restructure knowledge, resolve conflicts between KSs, infer new relationships, or generate new insights by combining information from different KSs or Clouds, guided by the current context or an active LoT. It's the mechanism that allows Clouds to be more than static containers.
    \item \textbf{Knowledge Lines (KLines) and Lines of Thought (LoTs):} KLines are sequences of interconnected KSs representing specific reasoning pathways or knowledge structures within a domain. LoTs extend this concept, acting as purpose-built, explicit paths that connect discrete knowledge nodes (KSs), potentially spanning multiple Clouds, Dimensions, or Junctures. LoTs explicitly represent logical progressions, decision-making sequences, diagnostic procedures, or query paths. They are crucial for guiding inference in a structured manner and serve as a fundamental mechanism for achieving explainability (XAI), providing a traceable audit trail of the reasoning process. For instance, an LoT could trace the steps taken to identify a potential threat in the naval scenario, linking sensor data interpretation (KS) to threat assessment criteria (KS) and recommended actions (KS), making the entire process transparent.
    \item \textbf{Other Key Concepts:} KERAIA also includes:
    \begin{itemize}

        \item \textbf{Forks:} Representing points within an LoT where reasoning paths diverge, allowing exploration of alternative hypotheses, scenarios, or decision branches.
        \item \textbf{Junctures:} Points where different LoTs or knowledge dimensions intersect or merge, allowing for the integration of information or reasoning from different perspectives.
        \item \textbf{Dimensions:} Used to structure knowledge multi-dimensionally, representing different perspectives (e.g., `TacticalView`, `LogisticsView`), layers of abstraction (`AbstractPlan`, `DetailedExecution`), or aspects of a problem (`SpatialDimension`, `TemporalDimension`).
        \item \textbf{Pulses/Impulses:} Mechanisms for triggering responders and initiating inference processes. Impulses often initiate LoTs or activate specific KSs, while Pulses propagate state changes or results.
        \item \textbf{Attractors:} Conditions within a KS that, when met, trigger specific responders or actions, enabling event-driven behavior.
        \item \textbf{Ultragraphs:} Interactive multimedia representations used during knowledge acquisition to capture complex relationships, processes, and expert insights visually, often serving as a precursor to formal KSYNTH representation.
        \item \textbf{Appellation/Convocation:} Appellation refers to the unique naming/identification system within KERAIA. Convocation refers to an assembly or collection of KSs working together, often within a Cloud or along an LoT.
    \end{itemize}
\end{itemize}

\begin{figure*}[ht]
    \centering
\includegraphics[width=0.8\linewidth]{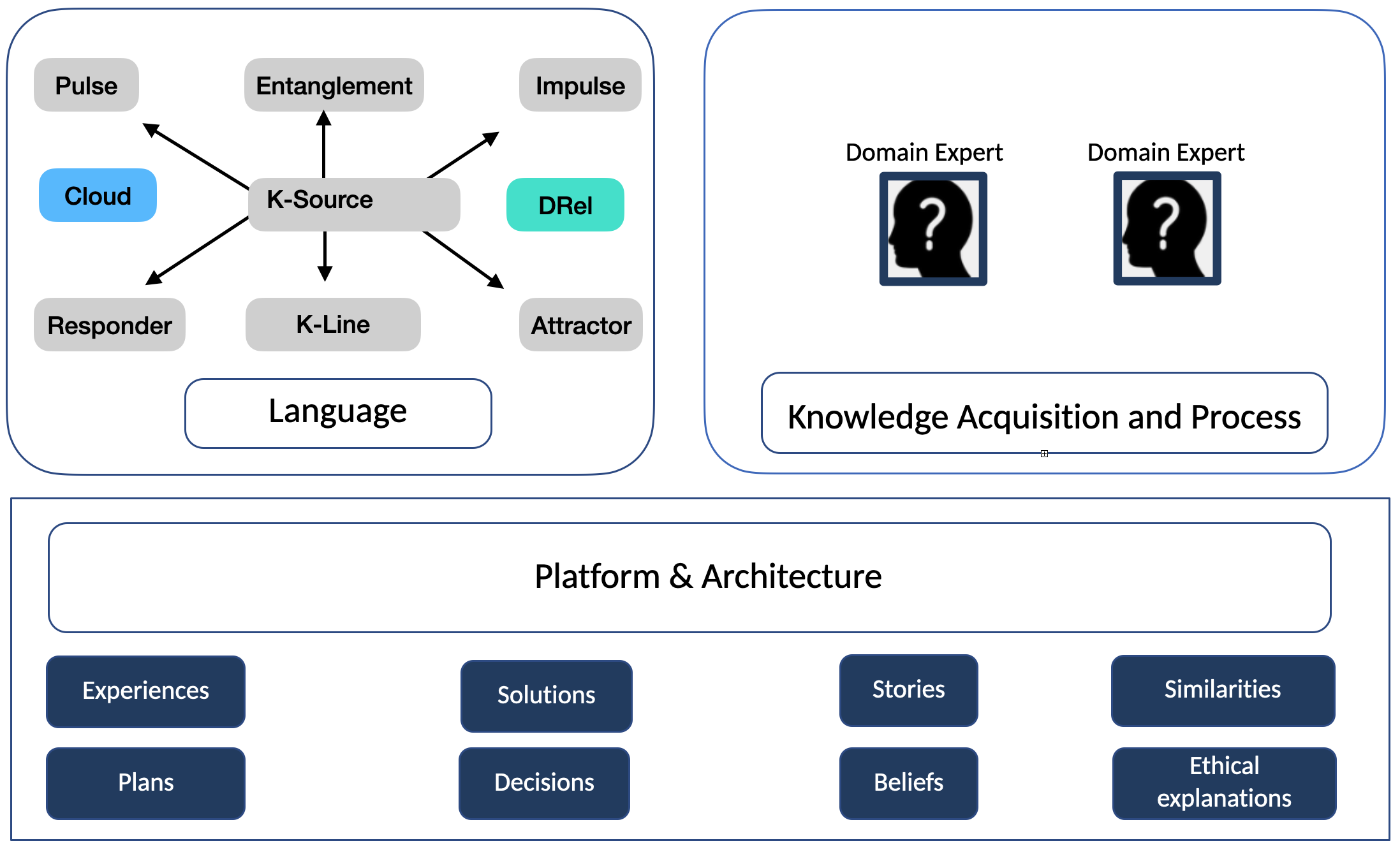}
    \caption{KERAIA - Functional overview}
    \label{fig:KERAIA - Functional overview}
\end{figure*}

\subsection{Architecture and Implementation}
\label{framework}

KERAIA is embodied in a software platform designed for knowledge engineering and real-time execution, aiming to bridge the gap between conceptual models and practical application. The architecture of the platform and the implementation details are crucial to understand its capabilities and feasibility.

\begin{figure*}[ht]
    \centering
    \includegraphics[width=0.8\linewidth]{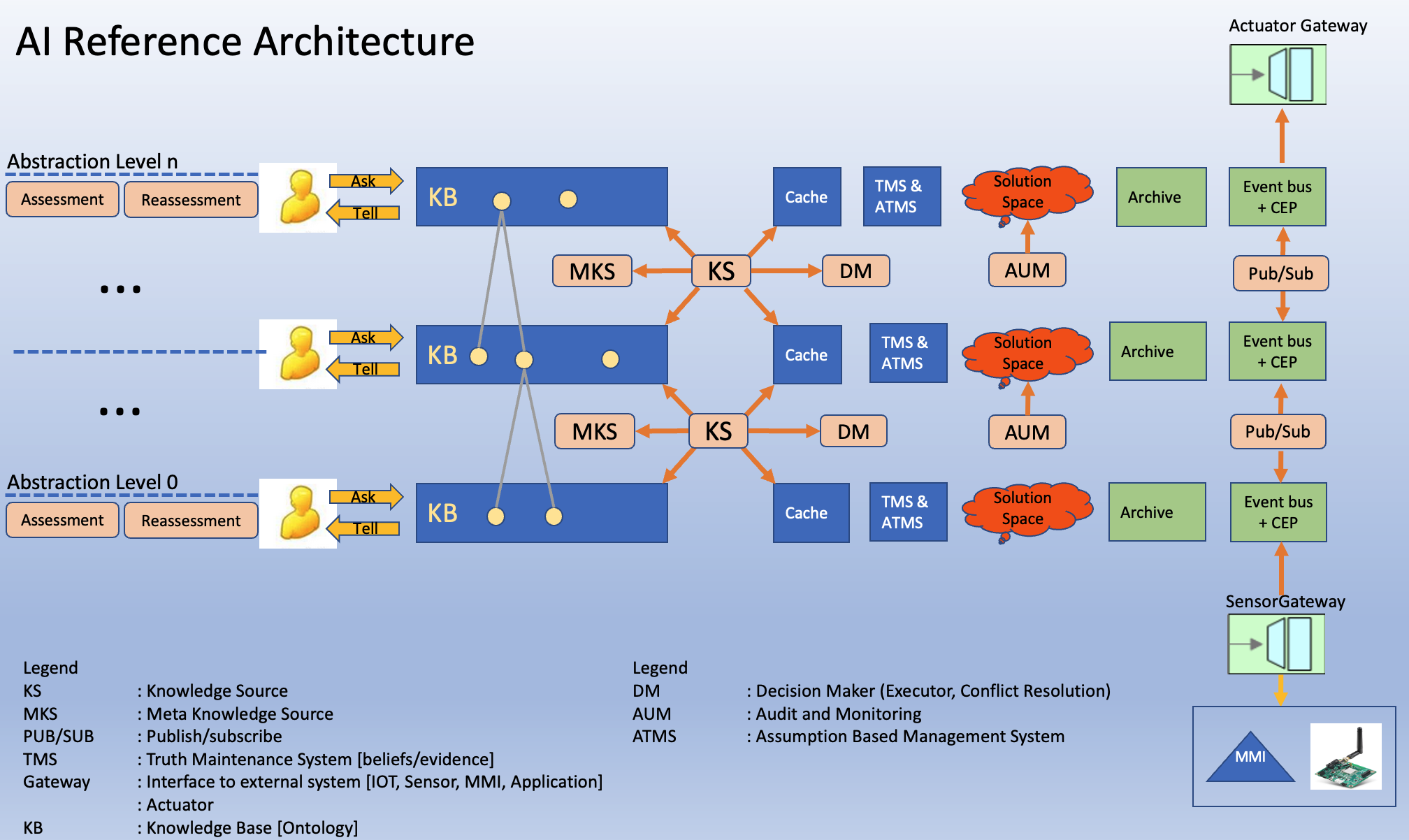}
    \caption{Architecture}
    \label{fig:Architecture}
\end{figure*}

\begin{itemize}
    \item \textbf{Platform Architecture Overview:} The architecture is conceptualized with distinct functional and component views. 
        \begin{itemize}
            \item Figure \ref{fig:KERAIA - Functional overview} details the software components, including a knowledge repository (storing KERAIA assets defined in KSYNTH), an execution platform, a messaging backbone (potentially Kafka for asynchronous communication, impulses / pulses, within convocations), and collaborative features (version control, token management for shared development). The execution platform supports a multilayered, multidimensional structure, enabling alternative world reasoning (e.g., exploring different scenarios via Forks in LoTs).
            \item Figure \ref{fig:Architecture} depicts the platform as both a knowledge repository and an execution environment supporting execution of multi-parameters, real-time, and across the enterprise. It highlights the flow from knowledge acquisition (using tools like web-based interviews and Q\&A templates) to the Knowledge Kernel, which forms the basis for generating solutions and insights.
            
        \end{itemize}
    \item \textbf{KSYNTH Language and Processor:} KERAIA employs a dedicated Knowledge Representation Language, KSYNTH. 
        \begin{itemize}
            \item \textit{Syntax and Structure:} KSYNTH extends the traditional frame syntax (slots, fillers) to allow arbitrary nesting depths, providing a structured text format for defining all KERAIA constructs: Clouds, KSs (with slots, responders, attractors), DRels (with conditions), LoTs (sequences of KSs), Dimensions, Forks, Junctures, etc. This deep nesting capability is argued to better represent complex, hierarchical, and interconnected knowledge compared to flatter structures. KSYNTH integrates various inference techniques, allowing for greater flexibility. The supported key paradigms include forward chaining, procedural reasoning, causal reasoning, explanation-based reasoning, reasoning by analogy, and anomaly detection. KSYNTH's strength lies in integrating these techniques within a unified syntax, enabling KERAIA to adapt to diverse knowledge engineering tasks by modeling each paradigm using knowledge sources (KSs) containing both knowledge and inference methods.
            \item \textit{Processor:} A KSYNTH processor is a core component of the platform, responsible for parsing these definitions, validating syntax against the language grammar, resolving references, and loading structured knowledge into the runtime environment for execution by the inference engine(s).
        \end{itemize}
    \item \textbf{Inference Integration (General Purpose Paradigm Builder - GPPB):} A key architectural innovation addressing the need for diverse reasoning capabilities is the GPPB. 

\begin{figure} [ht]
    \centering
    \includegraphics[width=0.8\linewidth]{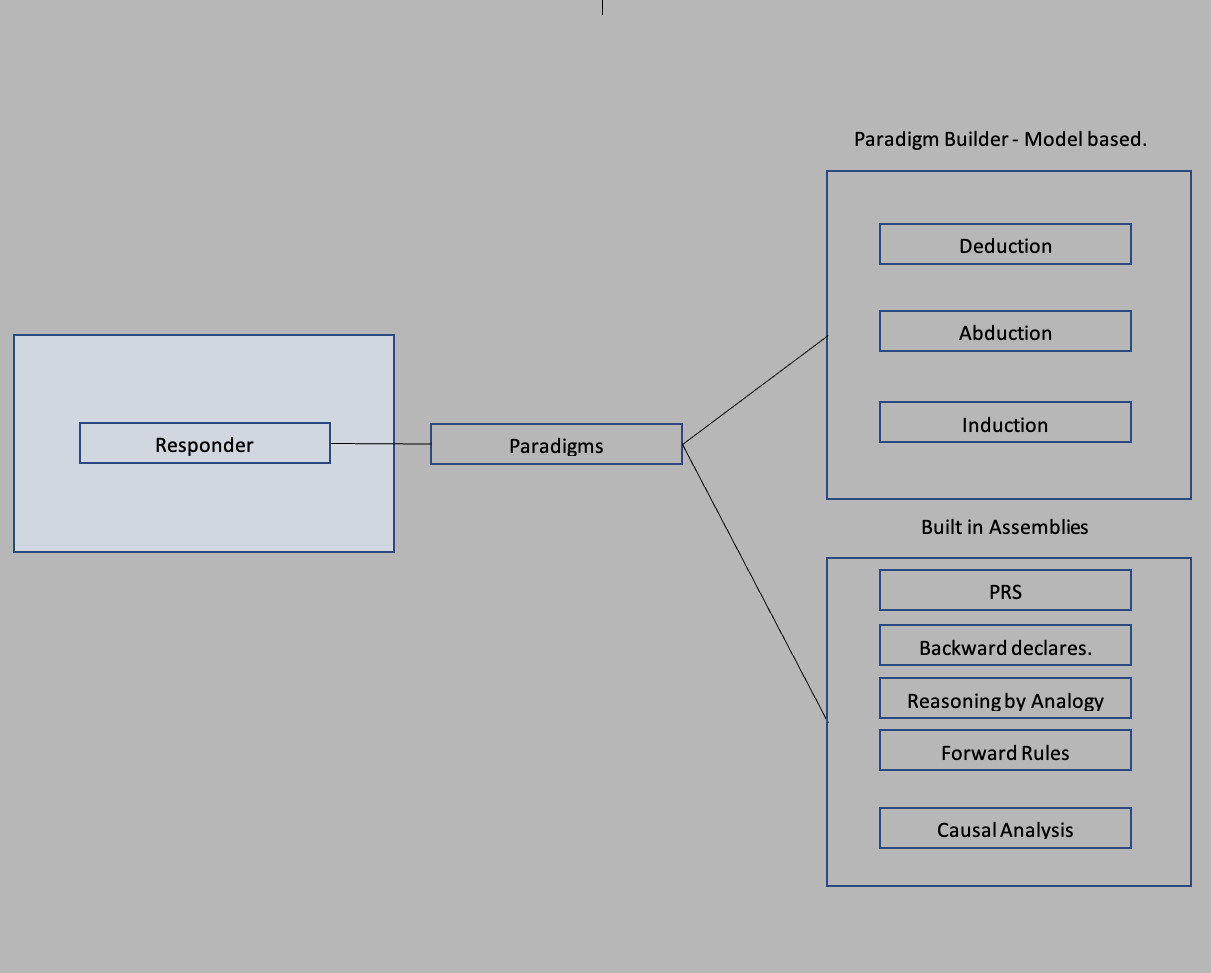}
    \caption{Assemblies}
    \label{fig:assemblies}
\end{figure}

        \begin{itemize}
            \item \textit{Multi-Paradigm Approach:} Instead of being limited to a single inference engine, KERAIA allows various inference techniques (procedural logic, causal reasoning, analogical mapping, rule-based inference, potentially calls to external ML models or simulators) to be associated with specific KSs via `responders`. 
            \item \textit{Integration Mechanism:} The GPPB provides tools, templates (referred to as 'Assemblies' in Figure \ref{fig:assemblies}), and interfaces for knowledge engineers or domain experts to define, configure, and integrate these domain-specific reasoning methods directly into the knowledge structure. It is a methodology designed for handling and transforming complex data structures (such as schema clouds that encompass frames and events) dynamically. It employs intricate templates, potentially resembling SPARQL constructs, that act as blueprints or metadata definitions to identify, extract, and structure data from the knowledge base. These input templates define the desired data structure and relationships, guiding the extraction process. Once defined, a template can be instantiated with specific instance values from the data set, tailoring operations to the context. Upon successful pattern matching and instantiation, the GPPB generates a new output structure based on the template, facilitating complex data transformations and the integration of various reasoning paradigms. Responders within KSs encapsulate the logic or calls to these GPPB-defined paradigms or external engines.
            \item \textit{Flexibility and Power:} This modular approach allows the system to leverage the most appropriate reasoning paradigm for different sub-problems within a larger task (e.g., using a rule engine for diagnostics, a procedural script for a standard operating procedure, a simulator for prediction). The GPPB acts as a bridge, allowing these diverse paradigms to interact coherently within the KERAIA execution environment, orchestrated by LoTs and KS activations.
        \end{itemize}
    \item \textbf{Knowledge Engineering Workflow and Tools:} The platform supports a defined workflow for knowledge engineering, with the goal of making the process more systematic and efficient.
        \begin{itemize}
            \item \textit{Workflow Steps:} The process typically involves: Knowledge Acquisition (e.g., using structured interviews with Q\&A templates  or potentially visual tools such as ultragraphs), Knowledge Representation (formalizing acquired knowledge in KSYNTH), Validation (checking consistency and correctness), Deployment (loading into the execution platform) and Refinement (iteratively improving the knowledge base based on performance and feedback).
            \item \textit{Supporting Tools:} KERAIA supports a web-based interface for visualization (schema graphs, hypergraphs, frame views - Figures \ref{fig:schemagraph}, \ref{fig:frameview}, \ref{fig:ultragraphs}) and interaction, suggesting that there are tools to aid engineers in building, navigating, and managing complex KERAIA knowledge bases. Usability enhancements are considered part of the practical implementation, aiming to lower the barrier for domain experts.
        \end{itemize}

 \begin{figure*} [ht]
    \centering
    \includegraphics[width=0.8\linewidth]{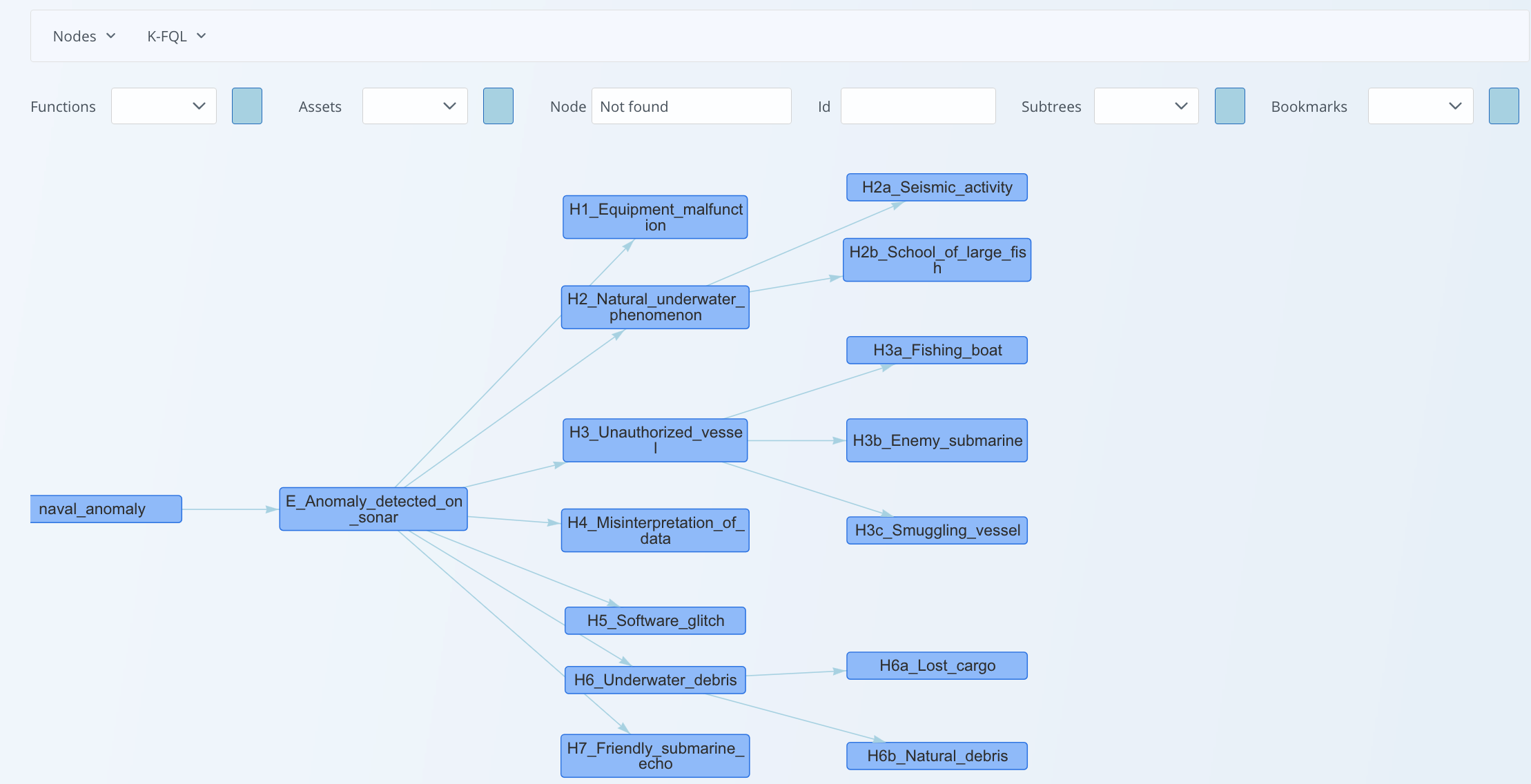}
    \caption{Schema graph}
    \label{fig:schemagraph}
\end{figure*}
\begin{figure*} [ht]
    \centering
    \includegraphics[width=0.8\linewidth]{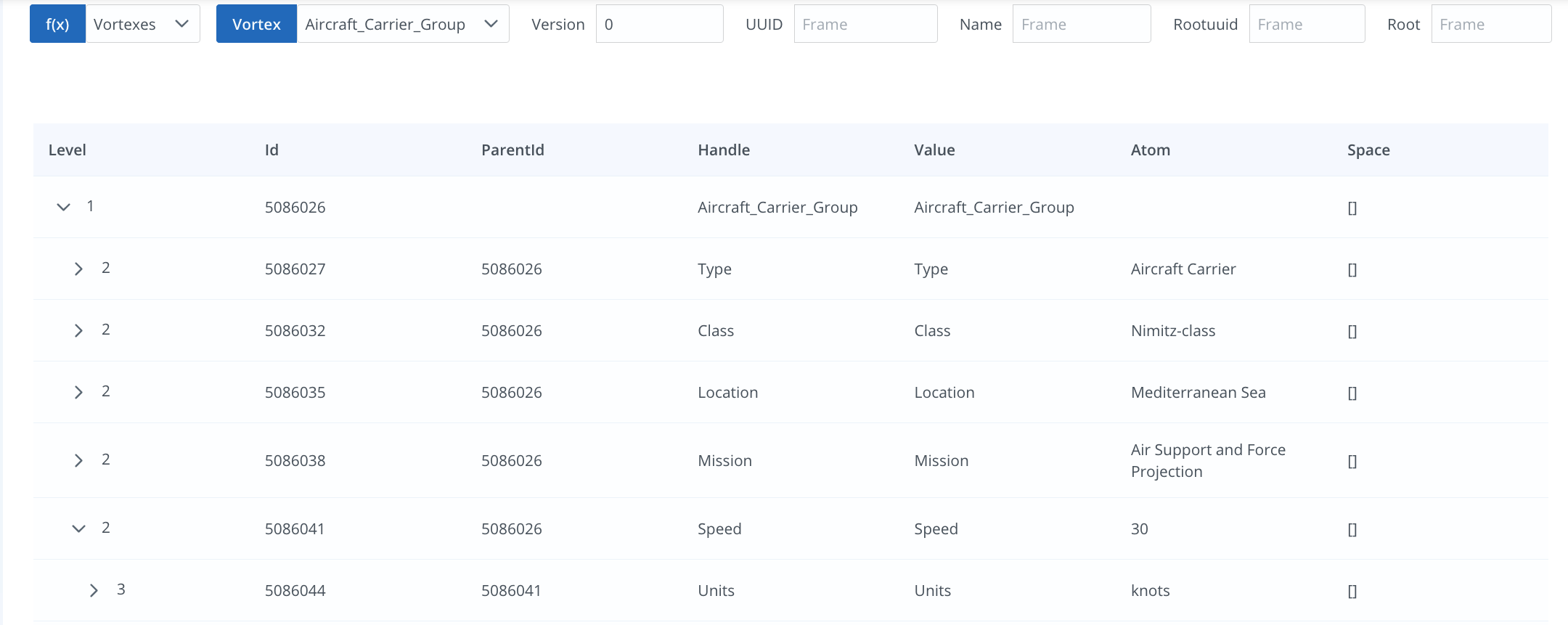}
    \caption{Frame View}
    \label{fig:frameview}
\end{figure*}
\begin{figure*} [ht]
    \centering
    \includegraphics[width=0.8\linewidth]{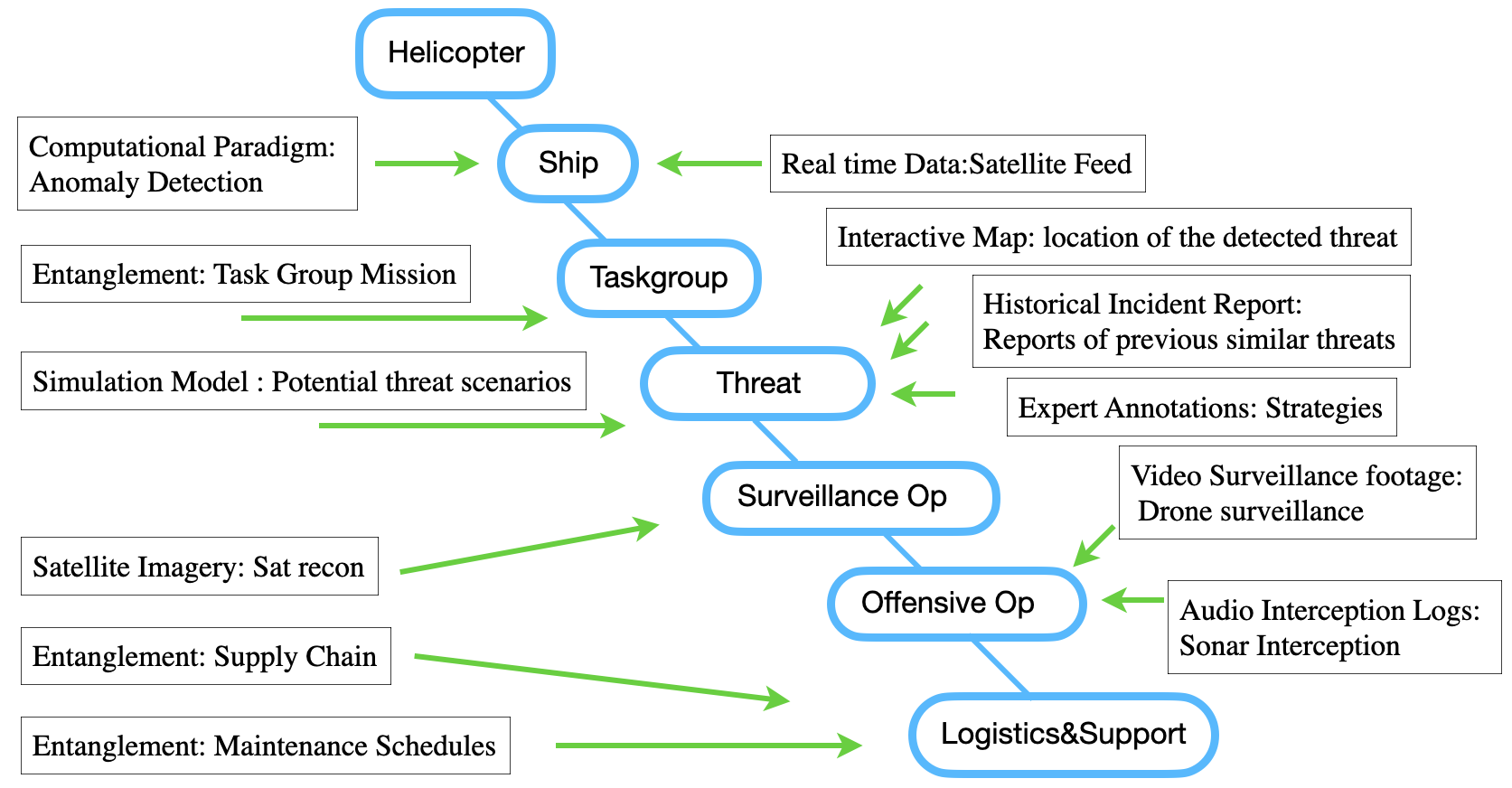}
    \caption{Ultragraphs}
    \label{fig:ultragraphs}
\end{figure*}

    \item \textbf{Implementation Details and Feasibility:} KERAIA is implemented as a robust software platform with supporting architectural diagrams. The described architecture (component-based, message-driven, web interface) aligns with modern software engineering practices. The existence of multiple detailed case studies (Water Treatment, Naval, RISK) successfully modeled and analyzed strongly implies that a functional implementation of the core framework (KSYNTH processor, execution engine capable of handling Clouds, KSs, DRels, LoTs, GPPB integration) exists and was used.The comparative analyses, including code volume comparisons, further support the notion of a concrete implementation. 
    Regarding computational feasibility : The dynamic nature of DRels and Cloud Elaboration inherently introduces computational overhead compared to static structures. Evaluating a DRel requires checking its conditions at run-time. Cloud elaboration might involve significant restructuring or inference. Performance depends heavily on the complexity of defined DRels/elaboration functions, the size of the knowledge base, and the update frequency. However, KERAIA's design attempts to manage this complexity:
        \begin{itemize}
            \item \textit{Locality:} Reasoning is often localized within KS responders or guided by specific LoTs, potentially limiting the scope of computation.
            \item \textit{Modularity (GPPB):} Allows optimizing specific inference paradigms independently.
            \item \textit{Contextual Activation:} Clouds and KSs may only be active or relevant within certain contexts, pruning the search space.
        \end{itemize}
    Although large-scale performance benchmarks are acknowledged as future work, the successful application to diverse, nontrivial case studies provides evidence that the approach is computationally feasible for complex symbolic reasoning tasks, particularly where its expressiveness, adaptability, and explainability offer significant advantages over potentially faster but more rigid or opaque alternatives. The trade-off between flexibility/expressiveness and raw computational speed is common in KR, and KERAIA appears viable for its intended application domains.
\end{itemize}

\subsection{Implementation Repository Overview}
\label{subgithub_overview}

The implementation details for the KERAIA (Knowledge Engineering and Reference AI Architecture) project are publicly available in a dedicated GitHub repository located at \url{https://github.com/srvarey/keraia}. This repository serves as a central hub containing both the source code and extensive supporting documentation that outlines the system's design, analysis, and application in various case studies, as well as alternative frameworks which were used for code volume comparison.

The core software implementation is located within the \texttt{impl} directory. The implementation is further organized into subdirectories corresponding to different reasoning paradigms illustrating how core constructs migh be implemented in other frameworks, These include modules for \texttt{case} (Case-Based Reasoning), \texttt{causal} reasoning, \texttt{kgraph} (Knowledge Graph operations), \texttt{onto} (Ontology management), and \texttt{rules} (Rule-Based Systems).

Beyond the executable code, the repository provides substantial documentation detailing the project's conceptual underpinnings and analytical work. The \texttt{analysis} directory houses numerous Markdown files covering domain-specific studies (e.g., naval carrier groups, pump diagnostics), conceptual modeling efforts (such as Clouds of Knowledge, Knowledge Paths, and fractal modeling approaches), interview notes, and comparative analyses such as a selection matrix of inference techniques. Similarly, the \texttt{ksynth} directory focuses on knowledge synthesis, offering documentation on mission execution logic, the application of rule-based systems (RBS) in missions, and a reference guide for a query language employed within the framework. Files related to Minsky's K-lines theory are found in the \texttt{klines} directory.

Rule definition and application are further detailed in dedicated sections. The \texttt{cypherrules} directory contains rules written in Cypher, the query language for the Neo4j graph database, indicating the use of graph databases for knowledge representation or reasoning. The \texttt{transformations} directory provides documentation, likely including pseudocode, on various rule-based transformation processes within KERAIA, covering general inference, Java-specific implementations, and transformations tailored for the naval scenario case study.

Insight into the user interaction aspect of KERAIA is provided by the \texttt{rulegui} directory. Although it contains no code, it includes screenshots (\texttt{facteditor.png}, \texttt{ruleeditor.png}) showcasing a graphical user interface (GUI) designed for editing the facts and rules utilized by the system. These images correspond to references found within the associated PhD thesis.

Specific applications of the KERAIA framework are documented in directories such as \texttt{ water treatment}. This section contains Markdown files detailing the water treatment plant case study, including troubleshooting procedures for issues like a blocked pump, transcripts of interviews, and examples illustrating the reasoning process applied in this scenario.

The \texttt{images} directory stores various visual assets used throughout the project documentation. Execution logs, such as \texttt{aip.log}, are found in the \texttt{logs} directory, potentially offering insights into runtime behavior. An \texttt{amendments.md} file suggests a mechanism for tracking changes or updates to the project or documentation.

In summary, the \texttt{srvarey/keraia} GitHub repository (\url{https://github.com/srvarey/keraia}) offers a comprehensive view of the KERAIA project. It combines core implementation code, primarily in Java and organized by reasoning paradigms, with extensive documentation covering analysis, design, knowledge modeling, rule systems, transformations, and specific case studies. This blend of code and documentation provides valuable details on the architecture's structure, functionality, and application.

\subsection{Explainability (XAI)}

KERAIA is designed with explainability as a core principle, not an afterthought. The primary mechanism for achieving this is the Line of Thought (LoT). By explicitly capturing the sequence of activated KSs and the reasoning steps performed (via responders) to reach a conclusion or decision, LoTs provide a transparent, step-by-step audit trail. Users or auditors can inspect an LoT to understand:
\begin{itemize}
    \item What knowledge sources (KS) were involved?
    \item In what sequence were they activated?
    \item What specific reasoning logic (responder) was executed at each step?
    \item How did the information flow between the KSs?
    \item Were there any alternative paths considered (via forks)?
\end{itemize}
This contrasts sharply with ‘black box’ systems (like many complex ML models) or even symbolic systems where inference chains are implicit and difficult to reconstruct in a human-understandable way. The KERAIA architecture also supports the generation of structured documentation of AI processes and decisions based on LoT execution, further enhancing transparency and trust. This inherent traceability is crucial for critical applications in defense, medicine, finance, and autonomous systems, where understanding ‘why’ a decision was made is often as important as the decision itself.

\subsection{Lines of Thought (LoTs) in Detail}

Lines of thought (LoT) are a pivotal construct in KERAIA, serving as the backbone of structured and explainable reasoning. Unlike implicit inference chains in some KR systems, LoTs are first-class explicit objects within the KERAIA framework. They represent directed pathways connecting sequences of Knowledge Sources (KSs), which could traverse multiple clouds, dimensions, and clusters.

\textbf{Purpose and Function:} LoTs serve multiple purposes:
\begin{itemize}
    \item \textbf{Guiding Inference:} They define specific reasoning sequences or problem solving strategies. By following an LoT, the system activates KSs in a predetermined order, applying their associated responders to process information or make decisions.
    \item \textbf{Structuring Complex Processes:} LoTs can model multi-step procedures, diagnostic workflows, or analytical processes, breaking down complex tasks into manageable, sequential steps.
    \item \textbf{Ensuring Explainability:} As each step in an LoT involves activating a specific KS and its responder, the entire reasoning path is explicitly recorded. This provides a clear, traceable explanation for how a conclusion was reached, fulfilling a key requirement of XAI.
    \item \textbf{Facilitating Knowledge Reuse:} Well-defined LoTs representing common reasoning patterns can be reused across different applications or scenarios.
\end{itemize}

\begin{figure*}[ht]
    \centering
    \includegraphics[width=0.7\linewidth]{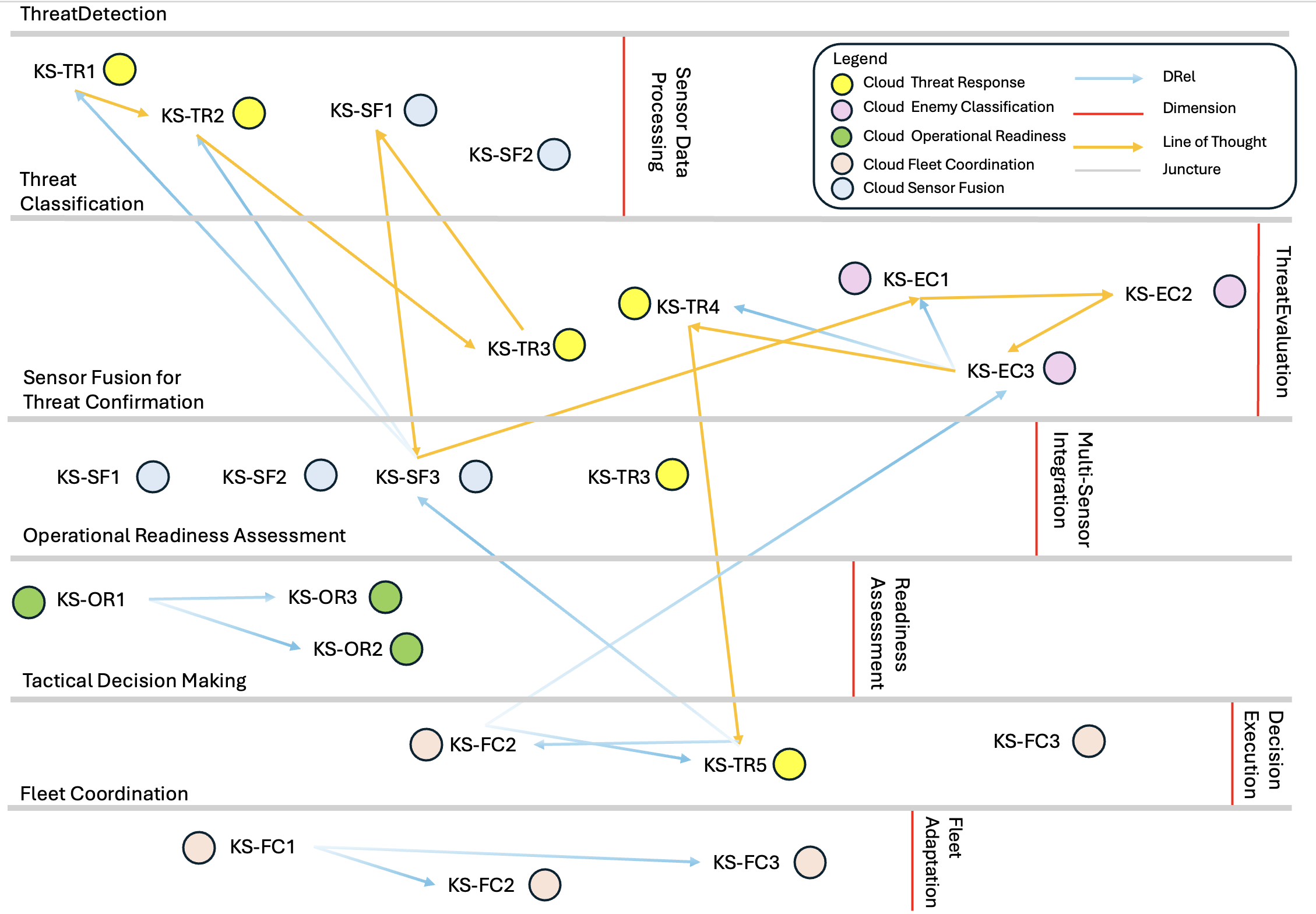}
    \caption{KERAIA Core Concepts.}
    \label{fig:coreconcepts}
\end{figure*}

\textbf{Example LoTs (Naval Scenario):} The naval warfare case study provides concrete examples of how LoTs structure the reasoning process. Referring to \ref{fig:coreconcepts}, we have the following:
\begin{itemize}
    \item \textbf{LoT-1 (Initial Radar Detection $\rightarrow$ Fusion Input):} 
        KS-TR1 (Radar System) $\rightarrow$ 
        KS-SF1 (Radar Data Buffer) $\rightarrow$ 
        KS-SF3 (AI Fusion System - Input Stage).
        *Purpose:* Capture raw radar data and prepare it for fusion.
    \item \textbf{LoT-2 (Initial Sonobuoy Detection $\rightarrow$ Fusion Input):} 
        KS-TR2 (Sonobuoy Network) $\rightarrow$ 
        KS-SF1 (Sonobuoy Data Buffer) $\rightarrow$ 
        KS-SF3 (AI Fusion System - Input Stage).
        *Purpose:* Capture raw sonobuoy data and prepare it for fusion.
    \item \textbf{LoT-3 (Data Fusion $\rightarrow$ Classification Input):} 
        KS-SF3 (AI Fusion System - Process Stage) $\rightarrow$ 
        KS-EC2 (Sensor Comparison Module) $\rightarrow$ 
        KS-EC3 (Classification System - Input Stage).
        *Purpose:* Fuse data from different sources and prepare the fused track for classification.
    \item \textbf{LoT-4 (Classification $\rightarrow$ Threat Assessment):} 
        KS-EC3 (Classification System - Process Stage) $\rightarrow$ 
        KS-EC1 (Intelligence Database Query) $\rightarrow$ 
        KS-FC2 (Threat Level Assessment Module).
        *Purpose:* Classify the track based on fused data and intelligence, then assess the threat level.
    \item \textbf{LoT-5 (Threat Assessment $\rightarrow$ Tactical Recommendation):} 
        KS-FC2 (Threat Level Assessment Module) $\rightarrow$ 
        KS-FC1 (Fleet Positioning Analysis) $\rightarrow$ 
        KS-FC3 (Tactical AI Recommender).
        *Purpose:* Analyze the threat in context of fleet status and generate tactical options.
    \item \textbf{LoT-6 (Recommendation $\rightarrow$ Command Decision Support):} 
        KS-FC3 (Tactical AI Recommender) $\rightarrow$ 
        KS-TR5 (Command Center Interface) $\rightarrow$ 
        [Human Decision Point / Action KS].
        *Purpose:* Present recommendations to the command for decision or automated action.
\end{itemize}

These examples demonstrate how LoTs provide structured, traceable pathways through the KERAIA knowledge base. Each step involves activating a specific KS, performing its associated reasoning or data processing (via responders), and potentially triggering the next KS in the sequence via impulses or other mechanisms defined within the LoT or KSs. By following these explicit paths, KERAIA ensures that its reasoning is not only effective but also transparent and understandable, directly addressing the requirements for Explainable AI and providing a clear answer to how specific conclusions are reached.

\textbf{Enhancing Explainability and Trust:} The explicit nature of LoTs is fundamental to KERAIA's commitment to Explainable AI (XAI). Understanding the reasoning process behind AI decisions is essential for establishing trust and transparency, particularly in critical domains where accountability is paramount. Traditional AI systems often lack the ability to provide clear and understandable audit trails for their reasoning. KERAIA addresses this by systematically capturing and presenting LoTs as clear, traceable pathways. This allows users to understand the rationale behind decisions, facilitating validation, and fostering trust in the system's outputs. The framework's focus on generating structured documentation and justifications for decisions further supports this goal, aiming to reduce operational risk and fulfill compliance requirements where necessary. Furthermore, explicit representation of LoTs opens possibilities for developing effective visualization techniques to make complex reasoning processes more accessible and understandable to human users, enhancing the human-AI collaboration potential.

\textbf{Core Explainability Features:} KERAIA incorporates several core features specifically designed to enhance transparency and explainability:
\begin{itemize}
    \item \textit{Lines of Thought (LoTs):} As detailed previously, LoTs provide explicit, traceable pathways for reasoning, forming the backbone of KERAIA's explainability.
    \item \textit{KS Explains:} Each Knowledge Source (KS) can contain an `explains` slot. This slot holds a human-readable justification for the KS's state, conclusion, or the action taken by its responders. These explanations can be static text or dynamically generated based on the KS's current context and the inputs it received. When an LoT is traversed, the `explains` content from each involved KS can be aggregated to form a comprehensive narrative of the reasoning process.
    \item \textit{Function Logs:} KERAIA logs the execution of functions, particularly those involved in Cloud Elaboration or KS responders. These logs capture input parameters, execution context, and output results, providing a detailed audit trail of computational steps.
    \item \textit{KS Versioning:} KERAIA supports versioning of KSs, allowing tracking of changes to knowledge representation over time. This is crucial for understanding how the knowledge base evolved and for debugging or auditing past reasoning processes.
    \item \textit{Narrative Analysis:} The logged explanations and function logs can be processed to generate coherent narratives that describe the reasoning of the system. This involves structuring the information chronologically or logically, filtering irrelevant details, and presenting it in a human-understandable format.
    \item \textit{What-If Analysis:} KERAIA's structure facilitates \"what-if\" analysis. By modifying specific KS attributes or altering conditions within LoT predicates (often at Forks), users can explore alternative scenarios and observe how the system's reasoning and conclusions change. This provides insight into the sensitivity of the system to different inputs or assumptions.
\end{itemize}
These features work together to provide multiple layers of explanation, from high-level reasoning paths (LoTs) to detailed justifications within individual knowledge components (KS Explains) and computational steps (Function Logs), directly supporting the goals of transparent and trustworthy AI.

\section{Evaluation and Case Studies}
\label{evaluation}

Evaluating a comprehensive knowledge representation framework like KERAIA requires assessing its expressiveness, adaptability, explainability, usability, and computational characteristics across diverse domains. Although formal large-scale benchmarking remains an area for future work, this section discusses evaluation considerations and presents insights gained from applying KERAIA to three distinct case studies, demonstrating its capabilities.

\subsection{Naval Warfare Case Study Revisited}

Addressing the complexities of modern naval warfare requires AI systems capable of sophisticated reasoning, mirroring aspects of human cognition. Early cognitive science work, such as Minsky's exploration of K-Lines \citep{minsky1980}, highlighted the non-linear, interconnected nature of human reasoning pathways. This contrasts with traditional, often layered, naval warfare models that focus on distinct stages such as data fusion, situation assessment, and resource allocation. While foundational, these earlier models face challenges with the increasing complexity and dynamism of modern multi-domain (including space and cyberspace) naval scenarios, which demand more adaptive and context-sensitive approaches to situation awareness and decision-making. KERAIA aims to address these challenges through its flexible knowledge representation and reasoning mechanisms.

The naval surveillance and threat response scenario, initially presented in \citep{varey2024keraia} and detailed extensively, serves as a primary validation of KERAIA's ability to handle complex, dynamic, multi-agent environments requiring sophisticated situation awareness and decision support. This case study was meticulously designed to showcase the interplay of KERAIA's core constructs in a realistic, high-stakes context.

\textbf{Scenario Context:} The scenario involves a naval task force operating in a potentially hostile area. The system must integrate data from various sensors (radar, sonar, ESM) to build a tactical picture, identify potential threats (e.g., incoming missiles, submarines), assess their intent, and recommend appropriate responses (e.g., evasive maneuvers, deploying countermeasures, engaging targets) while adhering to strict rules of engagement (ROE) and minimizing collateral damage. The dynamic nature involves changing threat postures, uncertain identifications, and the need for rapid adaptation. The importance of adaptability and structured reasoning in naval operations can be illustrated by considering mission scripts, such as those for platforms like the SH-60B Seahawk. Such scripts often have a dual nature: a static form providing a foundational operational plan, and a dynamic form that evolves based on real-time data and situational analysis. This dynamic adaptation is essential for handling the unpredictable scenarios encountered in naval warfare. KERAIA's KSYNTH scripting language and constructs like LoTs are designed to support this type of dynamic planning and execution. Furthermore, effective decision-making in complex naval scenarios benefits from structured cognitive frameworks. Cognitive models help break down intricate problems, organize knowledge, identify relationships, and provide a basis for decision-making. KERAIA facilitates the application of such models by providing structures (like Clouds, KSs, LoTs, Dimensions) that guide strategists through systematic problem-solving, ensuring relevant factors are considered and appropriate reasoning patterns are applied.

\subsection{KERAIA Implementation Details}

The implementation elaborates on the specific KERAIA structures used, emphasizing adaptability and context-sensitivity:
\begin{itemize}
    \item \textbf{Clouds of Knowledge (CoK) for Functional Decomposition and Context:} The problem was broken down using multiple, interconnected Clouds, representing functional areas and contextual layers. CoKs act as flexible, context-driven aggregations of Knowledge Sources (KSs), moving beyond rigid hierarchies (see \citep{varey2024keraia}, Section~4.2.1). They are designed to mirror the intricate, often non-linear nature of human cognition, allowing for a fluid aggregation where information can be dynamically linked and adapted. This is crucial in domains like naval reconnaissance where complex connections exist between intelligence tools, data sources, and algorithms, and relationships (e.g., alliances) can be multifaceted. Example Clouds included:
        \begin{itemize}
            \item \texttt{Cloud-SF} (Sensor Fusion): Correlating sensor tracks, managing track ambiguity.
            \item \texttt{Cloud-TR} (Threat Recognition): Classifying entities, assessing threats based on kinematics, signatures, and intelligence.
            \item \texttt{Cloud-FC} (Fleet Coordination): Resource allocation (weapon-target assignment), response planning, deconfliction.
            \item \texttt{Cloud-OR} (Operational Rules): Encapsulating ROE, doctrine, standard operating procedures.
            \item \texttt{Cloud-Intel}: Storing background intelligence (platform capabilities, known enemy tactics).
            \item \texttt{Cloud-Environment}: Representing environmental factors (weather, sea state) affecting sensors and operations.
        \end{itemize}
        This modularity allowed separation of concerns and mirrored the complex, interconnected nature of naval operations, enabling different reasoning processes to focus on relevant subsets of knowledge.
    \item \textbf{Knowledge Sources (KSs) for Entities and Processes:} Specific KSs represented entities (e.g., \texttt{KS-RadarTrack}, \texttt{KS-SonarContact}, \texttt{KS-FusedEntity}, \texttt{KS-OwnshipPlatform}, \texttt{KS-WeaponSystem}) and processing steps (e.g., \texttt{KS-TrackCorrelator}, \texttt{KS-ThreatAssessor}, \texttt{KS-ResponsePlanner}, \texttt{KS-ROEChecker}). Each KS contained relevant attributes (slots, e.g., position, velocity, classification, threat level, weapon status) and potentially responders (methods) implementing specific logic (e.g., a responder in \texttt{KS-ThreatAssessor} might calculate threat level based on kinematics and classification).
    \item \textbf{Lines of Thought (LoTs) for Reasoning Pathways:} Explicit LoTs defined the flow of reasoning, enhancing explainability . Examples included \texttt{LoT-SensorFusion}, \texttt{LoT-ThreatEvaluation}, and \texttt{LoT-ResponsePlanning}, connecting KSs across different Clouds. These LoTs explicitly mapped the sequence of analysis, from initial detection to recommended action, providing a traceable path for understanding system behavior.
    \item \textbf{Dynamic Relations (DRels) for Contextual Inheritance:} DRels were crucial for modeling context-dependent behavior. A key example involved a helicopter (\texttt{helo}) stationed on a ship (\texttt{ship}). A DRel allowed the \texttt{helo} KS to dynamically inherit the \texttt{speed} attribute from the \texttt{ship} KS *only when* the contextual condition \texttt{helo.location == ship} was true. When the helo took off, the inheritance ceased, demonstrating KERAIA's ability to model fluid real-world relationships dependent on dynamic context, unlike static \texttt{is-a} inheritance. Other DRels could model changing command relationships or sensor capabilities based on operational mode.
    \item \textbf{Forks for Decision Branching and Exploration:} Forks acted as conditional branching points within LoTs, allowing exploration of multiple reasoning paths based on predicates. For example, after risk assessment, a Fork could lead to different LoTs for immediate action (high threat, clear ID), further investigation (ambiguous track), or opportunistic analysis (low threat, potential intelligence target) based on the risk level and ROE constraints evaluated by a predicate function. Forks can connect different LoTs and may involve applying specific inference paradigms (like a forward-chaining cycle on ROE rules) to determine the branching condition.
    \item \textbf{Dimensions for Perspectives and Scenarios:} Dimensions allowed modeling different viewpoints or hypothetical scenarios. For instance, one Dimension could represent the tactical picture based purely on sensor data (`SensorDerivedDim`), while another (`OpportunityAnalysisDim`) could explore potential strategic opportunities, linked via a Fork. Another Dimension (`WorstCaseThreatDim`) could model a scenario assuming maximum hostile intent for ambiguous tracks, allowing for contingency planning.
    \item \textbf{Adaptive Inference Mechanisms (AIM) and Cloud Elaboration for Dynamic Adaptation:} AIM represents a suite of sophisticated functions that collectively constitute the \"elaborate\" process, enabling dynamic knowledge transformation within Clouds. This core operation transcends conventional hierarchical structures, offering a dynamic approach to knowledge evolution. It goes beyond simple instantiation, allowing complex transformations (e.g., applying inference paradigms via Forks) and the creation of \"clouds within clouds\". The AIM is not a monolithic entity but rather a suite of functions operating synergistically to analyze, transform, and synthesize knowledge. Each function addresses specific aspects, from simple categorization to complex integration of disparate knowledge clusters, forming the bedrock of the 'elaborate' operation. This allows AIM to accommodate a wide spectrum of elaborative processes, reflecting the intricate and recursive nature of human cognition where concepts can be delayed, altered, or expanded based on context. Through AIM, a cloud can dynamically adapt and reconfigure itself, ensuring the knowledge it contains is current, relevant, and richly interconnected.

    For example, in the naval context, a cloud that represents the strategic elements of a task force could be elaborated to incorporate a new conceptual layer for the readiness state. This readiness state cloud could dynamically encapsulate various operational clouds (logistics, armament status, personnel readiness), each potentially containing further sub-clouds. Similarly, a threat assessment cloud, initially containing broad categories, could be elaborated using AIM to include specific threat clouds (surface, subsurface, aerial), which could then be elaborated with details on capabilities, activities, and intentions. This dynamic evolution, driven by AIM, is crucial to maintaining accurate situation awareness.

    As new information arrived (e.g., updated track data, intelligence reports), Cloud Elaboration mechanisms, driven by AIM functions, could trigger updates to KS attributes, activate different responders, or alter LoT paths at Forks. 
    
    A detailed example illustrates elaborating an initial \texttt{Situation\_Element\_Perception\_Refinement} cloud into a more sophisticated \texttt{Strategic\_Situational\_Analysis} (or \texttt{Advanced\_Tactical\_Analysis}) cloud. This involved applying specific transformation functions, each with a generic type, to the initial KSs:
        \begin{itemize}
            \item To \texttt{Existence\_Size\_Analysis} KS (analyzes presence/size from radar/sonar):
                \begin{itemize}
                    \item \textbf{Function:} \texttt{Detailed\_Dimension\_Mapping} (Type: Augmentation). \textbf{Transformation:} Expands basic size data (e.g., overall dimensions) to include comprehensive dimensional data such as length, width, height and volume. \textbf{Output KS:} \texttt{Dimensional\_Profiles} (Provides 3D profiles crucial for detailed size profiling and operational planning).
                    \item \textbf{Function:} \texttt{Mass\_Estimation} (Type: Calculation). \textbf{Transformation:} Calculates the mass of objects based on their volumetric dimensions and assumed material densities. \textbf{Output KS:} \texttt{Mass\_Profiles} (Offers mass assessments vital to understanding logistical and kinetic properties, assessing threat levels, and planning responses).
                \end{itemize}
            \item To \texttt{Identity\_Analysis} KS (determines type, make, potential capabilities):
                \begin{itemize}
                    \item \textbf{Function:} \texttt{Capability\_Inference} (Type: Inference). \textbf{Transformation:} Infers technical and combat capabilities based on the type and class of the object, extracting possible functionalities (e.g. weapon systems, stealth features, sensor ranges). \textbf{Output KS:} \texttt{Capability\_Profiles} (Details possible capabilities essential for tactical assessments and strategic planning).
                    \item \textbf{Function:} \texttt{Operational\_Role\_Identification} (Type: Classification). \textbf{Transformation:} Classifies objects into potential operational roles such as surveillance, assault, support, or electronic warfare based on their identified capabilities and historical operational data. \textbf{Output KS:} \texttt{Operational\_Roles} (Helps in strategic mission planning, role assignment, and understanding object intent within potential conflict scenarios).
                \end{itemize}
            \item To \texttt{Kinematics\_Analysis} KS (evaluates motion characteristics: speed, trajectory, changes):
                \begin{itemize}
                    \item \textbf{Function:} \texttt{Predictive\_Trajectory\_Modeling} (Type: Prediction). \textbf{Transformation:} Models future trajectories of objects using their current kinematic data (speed, direction) along with environmental factors (currents, wind). \textbf{Output KS:} \texttt{Predictive\_Trajectories} (Aids in navigation planning, collision avoidance, engagement timing, and predicting future positions or intercept points).
                    \item \textbf{Function:} \texttt{Behavioral\_Pattern\_Recognition} (Type: Pattern Recognition). \textbf{Transformation:} Identifies and categorizes behavior patterns in movement that indicate specific tactical maneuvers (e.g., attack profiles, evasive actions, specific search patterns) or threat postures. \textbf{Output KS:} \texttt{Behavioral\_Insights} (Crucial for anticipatory defensive or offensive measures, understanding intent, and countering enemy tactics).
                \end{itemize}
        \end{itemize}
        This elaboration process, orchestrated by AIM, transforms raw or minimally processed perceptual data from the initial KSs into richer, strategically valuable insights contained within the new KSs of the \texttt{Strategic\_Situational\_Analysis} cloud. This deeper level of analysis, dynamically generated as needed, is key for operational planning and decision-making in complex naval scenarios, demonstrating how KERAIA can adaptively refine its understanding to maintain accurate situation awareness in a fluid environment.
\end{itemize}

\begin{figure}[ht] 
    \centering
    \includegraphics[width=0.9\linewidth]{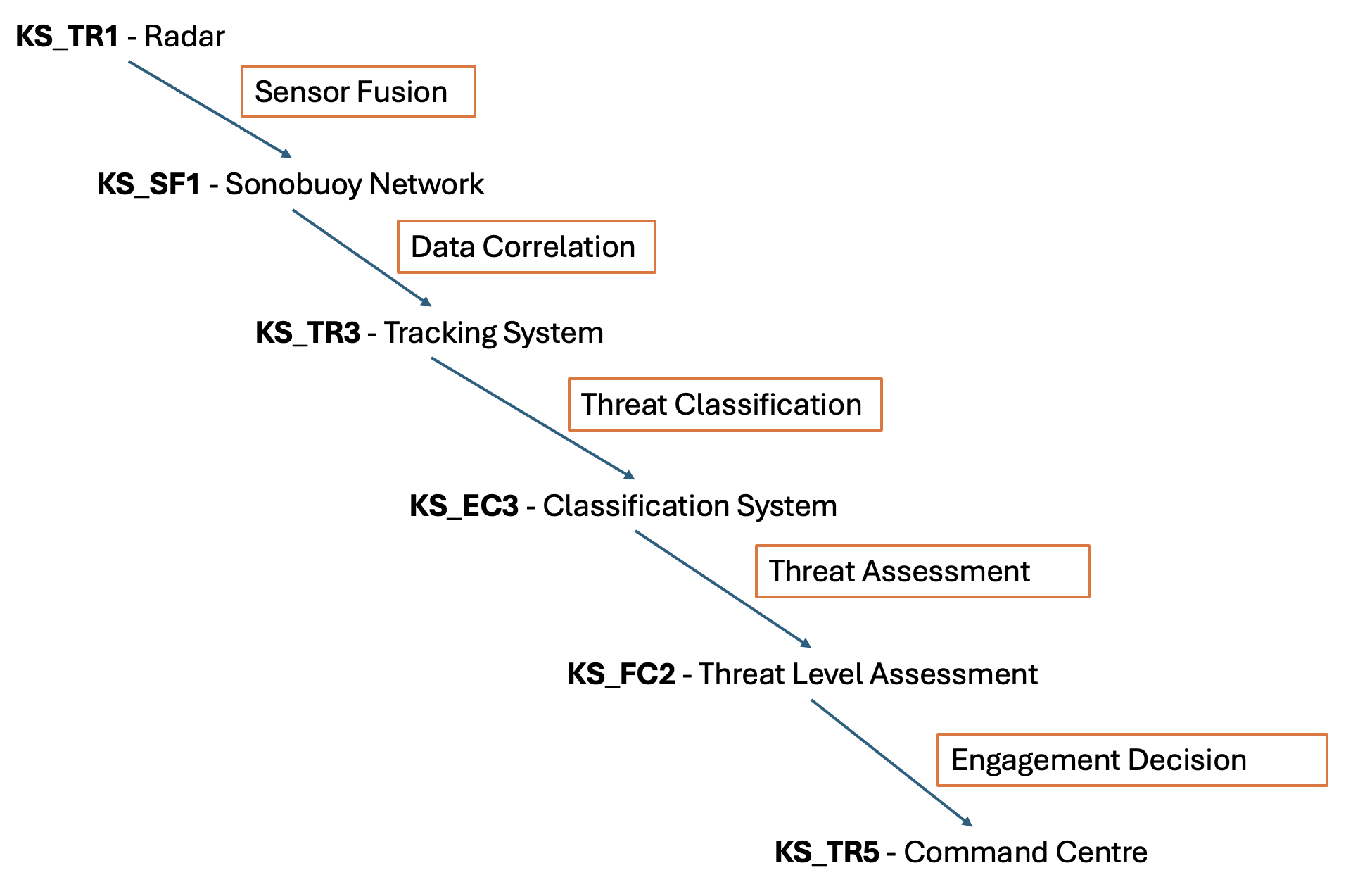}
    \caption{Naval Scenario Evolution . Illustrates the dynamic flow of information and reasoning steps within the KERAIA framework during the naval threat detection scenario, involving sensor fusion (Cloud-SF), threat recognition (Cloud-TR), and fleet coordination (Cloud-FC) clouds, guided by Lines of Thought.}
    \label{fig:Scenario_Evoltion_journal} 
\end{figure}

\textbf{Outcomes and Significance:} This expanded view of the naval case study demonstrates KERAIA's core strengths:
\begin{itemize}
    \item Representing complex, multi-agent systems with interacting components and processes using Clouds and KSs.
    \item Modeling dynamic behavior and context-sensitivity through DRels, AIM, and Cloud Elaboration. The detailed transformation functions (Detailed\_Dimension\_Mapping, Mass\_Estimation, Capability\_Inference, Operational\_Role\_Identification, Predictive\_Trajectory\_Modeling, Behavioral\_Pattern\_Recognition) exemplify how AIM facilitates the evolution from basic perception to strategic insight. For instance, `Detailed\_Dimension\_Mapping` provides the structural understanding needed for precise targeting or vulnerability analysis, while `Behavioral\_Pattern\_Recognition` allows the system to anticipate enemy actions based on movement, moving beyond simple tracking to infer intent. This elaboration process is crucial for maintaining accurate and relevant situation awareness in a fluid environment.
    \item Providing transparent reasoning pathways via explicit LoTs, enhancing explainability.
    \item Integrating diverse knowledge types (sensor data, intelligence, rules of engagement) within a unified framework.
\end{itemize}

\subsection{Implementation, Comparative Analysis, and Benchmarking Considerations}

As discussed in Section \ref{framework}, KERAIA exists as a functional software platform. The implementation details, while not exhaustively specified in terms of underlying technologies, support the core concepts. Comparative analysis suggests potential advantages in expressiveness and development effort for certain problems compared to combinations like Ontologies+SWRL or pure Rule-Based systems.

Regarding computational feasibility, the dynamic nature of DRels and Cloud Elaboration introduces overhead. Performance depends on the complexity of DRels/elaboration functions, knowledge base size, and update frequency. KERAIA manages this through:
\begin{itemize}
    \item \textit{Locality:} Reasoning is often localized within KS responders or guided by specific LoTs.
    \item \textit{Modularity (GPPB):} Allows optimizing specific inference paradigms.
    \item \textit{Contextual Activation:} Clouds and KSs may only be active within certain contexts.
\end{itemize}
Although large-scale performance benchmarks are future work, the successful application to diverse case studies provides evidence of computational feasibility for complex symbolic reasoning tasks where KERAIA's advantages in adaptability and explainability are paramount.

\textbf{Comparative Analysis :} A comparative analysis was performed against several established knowledge representation and reasoning paradigms, including Ontologies+SWRL, Case-Based Reasoning (CBR), Knowledge Graphs (KG), Causal Reasoning (CR), and Rule-Based Systems (RBS), focusing on compactness, expressiveness, and inference efficiency. KERAIA demonstrated advantages, particularly in compactness (measured by code volume required for a representative problem) and expressiveness, allowing for more concise representation of complex, dynamic relationships compared to the often verbose or rigid structures of other paradigms.

\textit{Code Volume Analysis:} To quantify compactness, the code volume required to represent the core aspects of the naval surveillance scenario was measured for KERAIA and compared against estimates for other paradigms. Code volume was defined based on normalized lines of code (NLOC), considering language-specific constructs and normalization weights (e.g., weighting declarative statements differently from procedural code). KERAIA required significantly less code volume (see Table \ref{tab:code_volume_comparison}) compared to equivalent representations in Ontologies+SWRL or RBS. This compactness is attributed to KERAIA's integrated constructs like DRels, Clouds, and the expressive KSYNTH language, which reduce redundancy and allow for higher-level abstractions.

\begin{table}[ht]
\centering
\small
\renewcommand{\arraystretch}{1.3}
\begin{tabularx}{\linewidth}{|>{\raggedright\arraybackslash}X|c|c|}
    \hline
    \makecell{\textbf{Paradigm Compared}\\\textbf{to KERAIA}} & 
    \textbf{Code Volume} & 
    \makecell{\textbf{Increase}\\\textbf{vs KERAIA}} \\
    \hline
    KERAIA                 & 27  & \textbf{Baseline} \\
    Ontologies + SWRL     & 51  & 89\%   \\
    Case-Based Reasoning  & 54  & 100\%  \\
    Knowledge Graphs      & 100 & 270\%  \\
    Causal Reasoning      & 48  & 78\%   \\
    Rule-Based Systems    & 66  & 144\%  \\
    \hline
\end{tabularx}
\caption{Total Code Volume for Naval Surveillance Representation with Increase Comparison}
\label{tab:code_volume_comparison}
\end{table}

\textit{Paradigm-Specific Comparisons:}
\begin{itemize}
    \item \textbf{Ontologies + SWRL:} Offer strong formal semantics and reasoning capabilities (e.g., consistency checking, classification). However, they can be verbose for representing complex procedural or dynamic knowledge. SWRL rules, while powerful, can become complex and computationally expensive, especially with frequent updates to the knowledge base. Representing the dynamic state changes and contextual adaptations in the naval scenario required extensive axiomatisations and rule sets, contributing to higher code volume compared to KERAIA's more integrated approach.
    \item \textbf{Case-Based Reasoning (CBR):} Excels at leveraging past experiences but struggles with case explosion, adaptation of solutions to novel situations, and generalization. Representing the vast state space and dynamic interactions of the naval scenario would require an impractically large case base or complex adaptation rules, making it less suitable than KERAIA for this type of problem.
    \item \textbf{Knowledge Graphs (KG):} Effective for representing interconnected entities and relationships. However, standard KGs often lack sophisticated inference mechanisms beyond path traversal or simple pattern matching. Representing complex rules, procedural knowledge, or dynamic relationships (like those captured by DRels) requires extensions or integration with other reasoning engines, potentially increasing complexity and code volume.
    \item \textbf{Causal Reasoning (CR):} Focuses on cause-and-effect relationships, useful for diagnostics and prediction. However, CR models often assume fixed dependencies and may struggle with the highly dynamic and context-dependent interactions present in naval warfare or complex process control. KERAIA integrates causal reasoning capabilities within its broader framework but is not limited to purely causal models.
    \item \textbf{Rule-Based Systems (RBS):} Widely used and well-understood. However, large RBS can suffer from rule explosion, difficulty managing rule interactions, and inference bottlenecks. Representing the multi-faceted strategies and dynamic context-switching required in the naval or RISK scenarios could lead to a very large and potentially brittle rule base compared to KERAIA's more structured and modular approach using Clouds, KSs, and LoTs.
\end{itemize}

In summary, KERAIA aims to mitigate the limitations of individual paradigms through its dynamic, multi-paradigm approach facilitated by KSYNTH and core constructs like Clouds, KSs, DRels, and LoTs. While trade-offs exist (e.g., potential complexity in knowledge engineering), the comparative analysis suggests advantages in compactness and expressiveness for complex, dynamic domains. The optimal paradigm choice ultimately depends on the specific application context and requirements.

\textbf{Benchmarking Considerations:} While comprehensive quantitative benchmarking is ongoing, a framework for evaluation, the Multidimensional Knowledge Elicitation and Validation Index (MKEVI), has been conceptualized. MKEVI proposes metrics beyond simple accuracy, aiming to capture the richness and robustness of knowledge-based systems like KERAIA. Key dimensions include:
\begin{itemize}
    \item \textit{Inference Depth Complexity (IDC):} Measures the sophistication of reasoning chains.
    \item \textit{Knowledge Breadth (KB):} Assesses the range of domains and contexts handled.
    \item \textit{Adaptability and Evolution (AE):} Evaluates adaptation to new information.
    \item \textit{Inference Efficacy (IE):} Considers correctness and speed of inference.
    \item \textit{Explanation and Justification (EJ):} Assesses the clarity and logical soundness of explanations (linking to LoTs).
    \item \textit{Collaborative Knowledge Building (CKB):} Measures support for collaborative development.
    \item \textit{Novelty and Innovation (NI):} Assesses generation of new insights.
    \item \textit{Multidimensional Paradigm Representation (MPR):} Evaluates proficiency in using diverse paradigms.
    \item \textit{Knowledge Structure Complexity (KSC):} Measures the richness and granularity of the knowledge base.
\end{itemize}
Applying such a multi-faceted benchmark is crucial for evaluating the true capabilities of advanced symbolic systems like KERAIA, going beyond traditional performance metrics to assess aspects like adaptability, explainability, and knowledge depth, which are central to its design philosophy.

\subsection{Generalizability: Water Treatment Case Study}

To demonstrate KERAIA's applicability beyond military simulations, a case study focused on diagnostics and decision support in a water treatment plant was developed. This domain presents different challenges, including reasoning about physical processes, sensor data interpretation, fault diagnosis, and adherence to operational procedures and safety regulations.

\textbf{Scenario Context:} The system monitors a water treatment plant, receiving data from sensors measuring parameters like flow rates, pressure, chemical levels (chlorine, pH), and turbidity. The goal is to detect anomalies, diagnose potential faults (e.g., pump failures, filter blockages, chemical dosing issues), predict consequences (e.g., impact on water quality), and recommend corrective actions to operators, ensuring safe and efficient operation.

\textbf{KERAIA Implementation Details:} The KERAIA model for this case study included:
\begin{itemize}
    \item \textbf{Clouds for Plant Sections and Processes:} Clouds represented major sections of the plant (e.g., `Cloud-Intake`, `Cloud-Filtration`, `Cloud-Disinfection`, `Cloud-Distribution`) and key processes (`Cloud-ChemicalDosing`, `Cloud-QualityMonitoring`).
    \item \textbf{KSs for Components and Parameters:} KSs modeled physical components (e.g., `KS-Pump`, `KS-Filter`, `KS-Valve`, `KS-Chlorinator`), sensors (`KS-FlowMeter`, `KS-PressureSensor`, `KS-TurbiditySensor`), and key parameters (`KS-WaterFlow`, `KS-ChlorineLevel`, `KS-WaterQuality`). Slots held current values, expected ranges, and status information.
    \item \textbf{Responders for Process Logic and Diagnostics:} Responders within KSs implemented logic for:
        \begin{itemize}
            \item Calculating derived values (e.g., total flow).
            \item Detecting deviations from normal operating ranges.
            \item Executing diagnostic rules (e.g., `IF PumpPressureLow AND PumpMotorCurrentHigh THEN Diagnose(PumpCavitation)`).
            \item Simulating process behavior (e.g., predicting downstream chlorine concentration based on dosing rate and flow).
        \end{itemize}
        The GPPB was leveraged to integrate potentially different types of diagnostic logic (e.g., rule-based, model-based) within relevant KS responders.
    \item \textbf{LoTs for Diagnostic Procedures and Response Protocols:} LoTs guided the diagnostic process. For example, an `LoT-HighTurbidityAlarm` might sequence through KSs to check filter status, backwash history, intake water quality, and chemical dosing effectiveness. Other LoTs could represent standard operating procedures for handling specific faults or initiating emergency shutdowns.
    \item \textbf{DRels for Component Interactions:} DRels modeled physical connections and dependencies. For instance, a `KS-Valve` could use a DRel to check the status of an upstream `KS-Pump` to determine expected inflow pressure, adapting its control logic based on the pump's operational state.
    \item \textbf{KSYNTH for Detailed Component Modeling:} The KSYNTH language allowed for detailed representation of component states and behaviors. For example, a `KS-Pump` might have nested slots for `MotorState` (On/Off/Tripped), `BearingTemperature`, `VibrationLevel`, and `EfficiencyCurve`. Responders could use this detailed information for more accurate diagnostics and predictive maintenance assessments.
\end{itemize}

\textbf{Outcomes and Significance:} This case study demonstrated KERAIA's ability to:
\begin{itemize}
    \item Model complex physical systems and processes.
    \item Integrate real-time sensor data with domain knowledge (component models, diagnostic rules, operational procedures).
    \item Implement structured diagnostic reasoning using LoTs.
    \item Adapt behavior based on component interactions and system state via DRels.
    \item Provide explainable diagnostic pathways through LoT traceability.
\end{itemize}
Successfully applying KERAIA to this industrial control domain provides strong evidence for its generalizability beyond the initial naval scenario, showcasing its potential for applications in process monitoring, fault diagnosis, and decision support in critical infrastructure.

\textbf{KSYNTH Scripting Language and KLines:} The KERAIA scripting language, KSYNTH, was central to modeling the water treatment scenario. KSYNTH provides a unified syntax for knowledge representation and integrating various inference techniques (Forward Chaining, Procedural Reasoning, Causal Reasoning, Explanation-Based Reasoning, Reasoning by Analogy, Anomaly Detection) alongside an assumption-based truth maintenance system (ATMS). A key feature utilized was KLines, which represent structured pathways through the knowledge base, linking related concepts and operational states. Unlike atomic facts in traditional rule-based systems, KLines offer significant advantages:
\begin{itemize}
    \item \textit{Contextual Relevance:} KLines embed context, ensuring retrieved data is relevant. For example, `WaterTreatmentSystem/WaterQuality/pH/CurrentValue` accesses pH within its specific application context.
    \item \textit{Efficient Navigation:} They provide direct paths, like file paths, reducing overhead compared to evaluating isolated facts.
    \item \textit{Reinforcement Learning:} Successful rule execution along a KLine can reinforce that pathway, increasing its priority for future similar scenarios.
    \item \textit{Analogical Reasoning Foundation:} The structured nature allows pattern recognition, enabling the system to apply successful pathways from one context to another, facilitating cross-contextual learning.
\end{itemize}

\textbf{Dimensions and Junctures for Contextual Reasoning:} To handle the temporal and contextual complexities of the water treatment plant, KERAIA's concepts of Dimensions and Junctures were employed. Junctures represent different levels of abstraction or viewpoints (e.g., `Juncture-WaterQualityManagement`, `Juncture-PumpOperation`), while Dimensions act as horizontal branches within each juncture, representing alternative scenarios, time frames, or perspectives. This structure allows the system to:
\begin{itemize}
    \item Model temporal changes and adapt reasoning over time.
    \item Explore different 'what-if' scenarios by reasoning across different dimensions.
    \item Make more comprehensive decisions by integrating multiple factors and perspectives, guided by assumptions and justifications embedded within dimensions.
    \item Align more closely with human reasoning, which naturally considers multiple viewpoints and adjusts based on new information.
\end{itemize}

\textbf{Unified Inference Integration:} KSYNTH facilitates the integration of multiple inference paradigms by using a unified format for input data sets (collections of clouds, KSs, and KLines). This standardization offers several benefits:
\begin{itemize}
    \item \textit{Interoperability:} Different reasoning engines (rule-based, procedural, causal, etc.) can seamlessly process the same input data.
    \item \textit{Reusability:} The same input set can be used across various paradigms for different tasks (e.g., diagnosis, anomaly detection).
    \item \textit{Maintainability \& Extensibility:} Adding new features or paradigms is simplified due to the consistent input structure.
    \item \textit{Composite Inferences:} Outputs from one paradigm can directly feed into another, enabling complex, multi-step reasoning chains.
    \item \textit{Adaptation:} Input sets can incorporate real-time data, allowing paradigms to adapt to dynamic conditions.
\end{itemize}
This unified approach allows KERAIA to leverage the strengths of different inference techniques effectively within the water treatment diagnostic system.

\subsection{Generalizability: RISK Board Game Case Study}

To further test KERAIA's capabilities in strategic planning and decision-making under uncertainty, a case study involving the classic board game RISK was developed. This domain requires reasoning about resource allocation, territory control, opponent modeling, and long-term strategy in a competitive, turn-based environment.

\textbf{Scenario Context:} The goal was to create an AI player for RISK using KERAIA to represent the game state, encode strategic heuristics, and make decisions during its turn (reinforcement, attack, fortification phases).

\textbf{KERAIA Implementation Details:} The implementation involved:
\begin{itemize}
    \item \textbf{Clouds for Game State and Strategy:} A primary `Cloud-GameState` held KSs representing the current board state (territories, ownership, army counts, continents, player status). Separate Clouds could potentially represent different strategic postures (e.g., `Cloud-AggressiveExpansion`, `Cloud-ContinentDefense`).
    \item \textbf{KSs for Game Elements:} KSs modeled `Territory`, `Continent`, `Player`, `Card`, etc., with relevant attributes.
    \item \textbf{Responders and Rules for Strategy:} Strategic heuristics, derived from expert human play, were encoded primarily as rules within KS responders or managed via the GPPB. Examples:
        \begin{itemize}
            \item Reinforcement: Prioritize placing armies to defend borders of owned continents, secure vulnerable territories, or prepare for attacks.
            \item Attack: Identify weakest adjacent enemy territories, calculate attack probabilities, prioritize attacks that complete continent ownership or eliminate weak players.
            \item Fortification: Move armies from secure interior territories to reinforce borders or consolidate forces after attacks.
        \end{itemize}
        These rules often involved querying the `Cloud-GameState` (e.g., finding adjacent territories, comparing army counts, checking continent ownership).
    \item \textbf{LoTs for Turn Phases:} LoTs structured the AI player's turn, sequencing through the Reinforcement, Attack, and Fortification phases, activating the relevant strategic rules/responders at each stage.
    \item \textbf{Integration with Game Simulator:} The KERAIA AI player interacted with an external RISK game simulator. 
        \begin{itemize}
            \item \textbf{State Updates:} The simulator published game state changes (e.g., after an opponent's move) via a message broker (e.g., Kafka). A KERAIA `DataFusion` process subscribed to these messages and updated the corresponding KSs within `Cloud-GameState`, potentially using KLines to deliver the frame-based representation from the simulator. This process ensured the AI's internal representation remained synchronized with the external simulation. The use of KLines for data delivery highlights their role in bridging structured representations between different system components.
            \item \textbf{Action Generation and Communication:} When rules fired based on the current game state in the KERAIA knowledge base, their consequents generated specific RISK game commands (e.g., \texttt{reinforce(territory, armies)}, \texttt{attack(fromTerritory, toTerritory, numDice)}, \texttt{fortify(fromTerritory, toTerritory, armies)}). These commands were then published onto the \texttt{risk} topic via the message broker, which the \texttt{GameAgent} simulator subscribed to, thereby executing the AI's chosen actions in the game environment. This closed the loop between the AI's reasoning and the game simulation, demonstrating a complete agent control cycle.
        \end{itemize}
    \item \textbf{Results and Generalizability Insights:} Simulations compared the KERAIA-based `AIAsset` bot (using different rule-based strategies like ``attack weakest'' or ``attack strongest'') against baseline bots (random, benevolent, cheater). The results showed continent ownership over game steps for different bot combinations. Key findings include:
        \begin{itemize}
            \item The KERAIA `AIAsset` consistently outperformed baseline bots, demonstrating the effectiveness of the encoded strategic knowledge.
            \item Different strategies encoded in KERAIA led to measurably different performance characteristics, highlighting the framework's ability to represent and execute distinct strategic approaches.
            \item The framework successfully managed the dynamic game state and executed turn-based decision cycles.
        \end{itemize}
\end{itemize}

\textbf{Representation Details:} The internal representation is handled by KSRL. The game board can be visualized as a hypergraph, with continents and territories as nodes. In a specific game state, this representation is adorned with army counts and ownership. KERAIA utilizes frames to represent game elements like countries, armies, and continents. This frame-based structure, which can be viewed hierarchically, allows for detailed representation, even for large knowledge graphs where visualizing localized information can be challenging. This contrasts with some graph implementations requiring complex queries to filter the knowledge space.

\textbf{Scenario Analysis and Rule Generation:} To facilitate rule construction, a feature vector approach, termed a \"threat table\", was used. This table lists friendly-occupied countries, neighboring countries (friend or foe), and border status relative to continents. This tactical view aids in creating rules for immediate actions. A complementary strategic view, listing countries ordered by continent, helps in formulating long-term goals. This structured analysis allows for translating expert heuristics into concrete rules. For example, deciding whether to attack the weakest or strongest border, or how to split forces when facing multiple threats, can be encoded based on this analysis. The knowledge acquisition process involved engaging with domain experts (experienced RISK players) to understand their strategies. This involved translating interview transcripts and scenario discussions into KLines, which were then transformed into knowledge sources and eventually into executable rules within the KERAIA framework. This process emphasizes capturing the expert's reasoning, including handling ambiguities and generating alternative strategies.

\textbf{Architecture and Integration:} The RISK use case instantiated a three-layered architecture (data fusion, situation assessment, resource allocation) utilizing KERAIA's multi-dimensional capabilities. A message broker (e.g., Kafka) facilitated loose coupling between the KERAIA AI (`AIAsset`) and the external `GameAgent` simulator. The `GameAgent` published game state events (e.g., board changes, phase transitions) to a `datafusion-post` topic. The KERAIA `DataFusion` process subscribed to this, updated its internal frame-based representation (delivered via KLines), and created working memory facts for a forward-chaining inference engine. Rules, partitioned by game phase (reinforce, attack, fortify), were triggered based on the game state. Fired rules generated game commands (e.g., `attack`, `fortify`), which were published to a `risk` topic, consumed by the `GameAgent` to execute the AI's actions, thus closing the control loop.




\textbf{Rule Implementation and Interaction Details:} Further insight into the practical implementation of the rule-based logic within the `AIAsset` is available. The partitioning of rules based on the game phase (Reinforcement, Attack, Fortification) was managed by the `DataFusion` process. When the `GameAgent` simulator published an event indicating a phase change (e.g., `PhaseChange(Attack)`), the `DataFusion` process updated the `GamePhaseCloud` KS. This change acted as a trigger, potentially managed via KERAIA's responder mechanism or simple conditional logic within the `DataFusion` process, to activate the appropriate rule set within the forward-chaining engine. For instance, upon entering the Attack phase, the Reinforcement rules would be deactivated (or ignored), and the Attack rules would become the primary focus for the inference engine.

Specific rule examples illustrate the translation of heuristics:
\begin{itemize}
    \item \textit{Reinforcement - Continent Security:}
\begin{lstlisting}[basicstyle=\ttfamily\small,breaklines=true]
IF Continent(ID=Australia).currentOwner == Self
   AND Territory(ID=Siam).owner == Self
   AND Territory(ID=Siam).armyCount < 5  % Example threshold
THEN GenerateReinforceAction(Territory=Siam, Armies=CalculateNeeded(Siam)) % CalculateNeeded could be a function call
END RULE
\end{lstlisting} (Conceptual KSYNTH-like rule)
\begin{lstlisting}[basicstyle=\ttfamily\small,breaklines=true]
FIND Territory T1 WHERE T1.owner == Self AND T1.armyCount > 1
FIND Territory T2 WHERE T2.owner != Self AND IsAdjacent(T1, T2)
MINIMIZE T2.armyCount AS MinEnemyArmies
WHERE T1.armyCount > MinEnemyArmies + 1 % Ensure sufficient advantage
THEN GenerateAttackAction(From=T1, To=T2, Dice=CalculateDice(T1.armyCount)) % CalculateDice determines optimal dice number
END RULE
\end{lstlisting}
    \item \textit{Fortification - Consolidate Border:}
\begin{lstlisting}[basicstyle=\ttfamily\small,breaklines=true]
FIND Territory T_Border WHERE T_Border.owner == Self AND IsBorderTerritory(T_Border)
FIND Territory T_Interior WHERE T_Interior.owner == Self AND IsConnected(T_Interior, T_Border) AND NOT IsBorderTerritory(T_Interior) AND T_Interior.armyCount > 1
MAXIMIZE T_Interior.armyCount AS MaxInteriorArmies
THEN GenerateFortifyAction(From=T_Interior, To=T_Border, Armies=T_Interior.armyCount - 1)
END RULE
\end{lstlisting}
\end{itemize}
These examples show how KSYNTH rules could combine state checks (ownership, army counts, adjacency retrieved via KS queries and potentially DRels for connectivity) with calculations or function calls (e.g., `CalculateNeeded`, `CalculateDice`, `IsBorderTerritory`) to implement nuanced strategies. The GPPB's role here is implicit in managing the execution context for these rules and potentially integrating external functions or algorithms called within the rule consequents.

The ability to easily swap entire rule sets corresponding to different high-level strategies (e.g., swapping the `AttackWeakestNeighbor` set for an `AttackStrongestChokepoint` set) was a key aspect demonstrated, allowing for rapid experimentation and evaluation of different strategic approaches within the simulation environment, directly contributing to understanding the impact of different knowledge encodings on agent performance.

\subsection{Summary of Evaluation}

The three case studies – naval warfare, water treatment diagnostics, and the RISK game – collectively provide strong qualitative evidence for KERAIA's expressiveness, adaptability, and generalizability across diverse and complex domains. They demonstrate the practical utility of core constructs like Clouds (for modularity and context), DRels (for dynamic relationships), and LoTs (for structured, explainable reasoning). The successful application to both real-world inspired problems (naval, water treatment) and strategic planning scenarios (RISK) directly addresses  concerns about the framework's applicability beyond its initial domain. While quantitative performance benchmarking remains essential future work, these detailed qualitative evaluations validate KERAIA's potential as a powerful and versatile framework for advanced symbolic knowledge engineering.

\section{Conclusions and Discussion}
\label{discussion}

The KERAIA framework, as detailed and evaluated through various case studies \citep{varey2024keraia}, presents a distinct approach to knowledge representation and reasoning, offering potential advantages while also having limitations and prompting avenues for future research.

\subsection{Advantages and Implications}

\begin{itemize}
    \item \textbf{Enhanced Adaptability and Context Sensitivity:} KERAIA's core strength lies in its ability to handle dynamic environments. Clouds of Knowledge, Cloud Elaboration, and especially Dynamic Relations (DRels) allow knowledge structures and inheritance patterns to adapt based on the current situation, moving beyond the often rigid nature of traditional ontologies or rule sets. This is crucial for real-world applications where context dictates relevance and behavior.
    \item \textbf{Improved Expressiveness for Complex Mental Models:} The framework, particularly the KSYNTH language and the concept of nested Clouds and multi-dimensional reasoning, aims to provide a richer language for capturing the complexity of human expert reasoning, including handling multiple perspectives, hypothetical scenarios (via Forks), and intricate dependencies.
    \item \textbf{Inherent Explainability (XAI):} Lines of Thought (LoTs) provide a built-in mechanism for tracing reasoning steps, significantly enhancing transparency and trustworthiness compared to black-box systems or KR paradigms where inference paths are implicit or difficult to reconstruct. This directly addresses a critical requirement in many domains.
    \item \textbf{Integration of Multiple Inference Paradigms:} The General Purpose Paradigm Builder (GPPB) concept allows domain-specific reasoning methods to be encapsulated within Knowledge Sources, enabling KERAIA to potentially leverage diverse inference techniques (procedural, causal, analogical, etc.) within a single, coherent framework, rather than being tied to one specific engine.
    \item \textbf{Facilitation of Knowledge Engineering:} The platform includes tools aimed at supporting collaborative knowledge acquisition and management, potentially streamlining the process of translating expert knowledge into executable AI systems.
\end{itemize}

These features suggest KERAIA could be particularly beneficial in domains like autonomous systems control, advanced diagnostics, intelligence analysis, and complex simulation environments where dynamism, context, and explainability are key challenges for traditional AI approaches.

\subsection{Limitations and Challenges}

Despite its potential, KERAIA also faces limitations and challenges:

\begin{itemize}
    \item \textbf{Scalability and Performance:} Rigorous quantitative evaluation of KERAIA's performance and scalability is needed. The overhead associated with dynamic evaluation of DRels, Cloud Elaboration, and managing complex LoTs could potentially impact performance in very large-scale or extremely time-critical applications. Further optimization and benchmarking are required.
    \item \textbf{Complexity of Knowledge Engineering:} While KERAIA aims to facilitate knowledge acquisition, the richness of its constructs (Clouds, DRels, LoTs, Dimensions, etc.) might introduce a steeper learning curve for knowledge engineers compared to simpler KR formalisms. Developing effective methodologies and intuitive tools for building and maintaining complex KERAIA knowledge bases will be crucial for practical adoption.
    \item \textbf{Formal Semantics and Verification:} While inspired by frames and logic, the formal semantics of KERAIA, particularly constructs like DRels and Cloud Elaboration, may require further development and clarification compared to well-established formalisms like Description Logics (used in OWL). This could impact formal verification and validation of KERAIA-based systems.
    \item \textbf{Integration with Sub-Symbolic AI:} KERAIA is primarily a symbolic framework. While the GPPB allows integration of diverse paradigms, seamless and principled integration with data-driven, sub-symbolic methods (e.g., deep learning) for tasks like perception or learning from raw data remains an open research area.
    \item \textbf{Maturity and Tooling:} Compared to mature technologies like OWL/RDF ecosystems or established rule engines, KERAIA is a newer framework. The robustness, feature completeness, and community support for its software platform and tools will influence its broader adoption.
\end{itemize}

\subsection{Future Work}

Addressing the limitations and building upon KERAIA's strengths suggests several directions for future research:

\begin{itemize}
    \item Conduct comprehensive performance benchmarking and scalability analysis.
    \item Refine knowledge engineering methodologies and tools for KERAIA.
    \item Further develop the formal semantics of KERAIA constructs.
    \item Explore deeper integration with machine learning and sub-symbolic methods.
    \item Apply KERAIA to a wider range of complex, real-world problems.
    \item Investigate automated methods for learning or refining KERAIA structures (Clouds, LoTs, DRels) from data or expert interaction.
\end{itemize}

In conclusion, KERAIA offers a promising direction for advancing symbolic AI, particularly in creating more adaptive, context-aware, and explainable systems. Its novel constructs provide powerful tools for modeling complex domains, although further research and development are needed to fully realize its potential and address challenges related to performance, complexity, and formalization.

\section{Conclusion}
\label{conclusion}

This paper has presented KERAIA, a novel knowledge engineering framework designed to overcome limitations of traditional KR paradigms in dynamic and complex environments. By extending frame-based concepts with innovative constructs such as Clouds of Knowledge, Dynamic Relations (DRels), Lines of Thought (LoTs), and Cloud Elaboration, KERAIA offers enhanced capabilities for adaptive knowledge representation, context-sensitive reasoning, and inherently explainable decision-making.

The architecture of the KERAIA platform, including the KSYNTH language and the General Purpose Paradigm Builder (GPPB), provides an integrated environment for knowledge acquisition and multi-paradigm inference. The framework's effectiveness and versatility were demonstrated through detailed case studies spanning naval warfare, water treatment diagnostics, and strategic gaming. These applications highlighted KERAIA's ability to model multi-layered situational awareness, fuse diverse intelligence sources, adapt reasoning pathways in real-time, and provide transparent reasoning traces via LoTs.

Compared to established methods like static ontologies (RDF/OWL), rule-based systems, or standard knowledge graphs, KERAIA provides distinct advantages in flexibility, context-sensitivity, and built-in explainability, making it well-suited for domains where these factors are critical. While challenges remain, particularly concerning quantitative performance evaluation and the complexity of knowledge engineering, KERAIA represents a significant step towards realizing more robust, adaptive, and trustworthy symbolic AI systems capable of tackling real-world complexity.

Future work will focus on performance optimization, refining the knowledge engineering process, further developing formal underpinnings, and exploring integration with sub-symbolic AI, aiming to solidify KERAIA's position as a powerful tool for next-generation intelligent systems.

\section{Acknowledgment}
The Anh Han is supported by EPSRC (grant EP/Y00857X/1).
\bibliographystyle{IEEEtranN} 
\bibliography{references}

\end{document}